\pdfoutput=1
\documentclass[11pt]{article}
\usepackage[final]{mtsummit25}
\usepackage{times}
\usepackage{latexsym}
\usepackage[T1]{fontenc}
\usepackage[utf8]{inputenc}
\usepackage{microtype}
\usepackage{inconsolata}
\usepackage{graphicx}
\usepackage{caption}
\usepackage{subcaption}
\usepackage{amsmath,amsfonts,amssymb}
\usepackage{textcomp}
\usepackage{url}
\usepackage{booktabs}
\usepackage{multirow}
\usepackage{arydshln}
\usepackage{makecell}
\usepackage{multirow}
\usepackage{pifont}
\usepackage[linesnumbered,ruled,vlined]{algorithm2e}
\usepackage{color}
\usepackage{copyright}

\title{Languages Transferred Within the Encoder:\\
On Representation Transfer in Zero-Shot Multilingual Translation}

\author{
\textbf{Zhi Qu\thanks{This work was done during the first author’s internship at Advanced Speech Translation Research and Development Promotion Center, National Institute of Information and Communications Technology, Kyoto, Japan.}$^\dag$ } \quad \textbf{Chenchen Ding$^\dag$$^\ddagger$} \quad
    \textbf{Taro Watanabe$^\dag$}
  \\ \ \\
  $^\dag$Nara Institute of Science and Technology, Japan
  \\
  \texttt{\{qu.zhi.pv5, taro\}@is.naist.jp}
  \\ \ \\
  $^\ddagger$National Institute of Information and Communications Technology, Japan
  \\
  \texttt{chenchen.ding@nict.go.jp}
}

\begin{document}
\maketitle
\begin{abstract}
Understanding representation transfer in multilingual neural machine translation (MNMT) can reveal the reason for the zero-shot translation deficiency.
In this work, we systematically analyze the representational issue of MNMT models.
We first introduce the identity pair, translating a sentence to itself, to address the lack of the base measure in multilingual investigations, as the identity pair can reflect the representation of a language within the model.
Then, we demonstrate that the encoder transfers the source language to the representational subspace of the target language instead of the language-agnostic state.
Thus, the zero-shot translation deficiency arises because the representation of a translation is entangled with other languages and not transferred to the target language effectively.
Based on our findings, we propose two methods: 1) low-rank language-specific embedding at the encoder, and 2) language-specific contrastive learning of the representation at the decoder.
The experimental results on Europarl-15, TED-19, and OPUS-100 datasets show that our methods substantially enhance the performance of zero-shot translations without sacrifices in supervised directions by improving language transfer capacity, thereby providing practical evidence to support our conclusions.
Codes are available at \url{https://github.com/zhiqu22/ZeroTrans}.
\end{abstract}

\section{Introduction}
State-of-the-art neural machine translation systems are adaptable to multilingualism, resulting in a single encoder-decoder model that executes arbitrary translations by adding a tag specified to the target language at the beginning of source sentence \cite{ZeroMNMT2016, GooglesMNMT,TagMatter_2021}.
Multilingual neural machine translation (MNMT) is theoretically attractive because zero-shot translations, i.e., translations unseen in training, allow the training of a multilingual model with minimal cost.
Unfortunately, the performance of zero-shot translations always lags behind \cite{aharoni-2019, arivazhagan-2019, gu-2019,TLP-2021, Constras-2021, target-off}.
\begin{figure}[t]
    \centering
      \begin{subfigure}[b]{0.49\linewidth}
        \centering
        \includegraphics[width=0.9\linewidth]{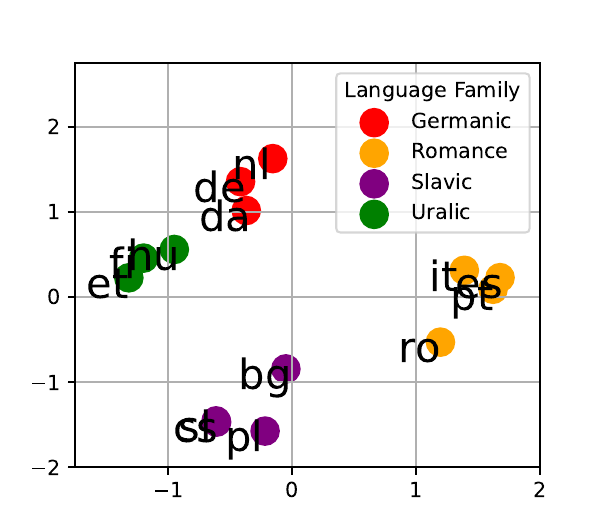}
        \caption{Cluster by target languages}
        \label{fig:cluster}
      \end{subfigure}
      \begin{subfigure}[b]{0.49\linewidth}
        \centering
        \includegraphics[width=0.99\linewidth]{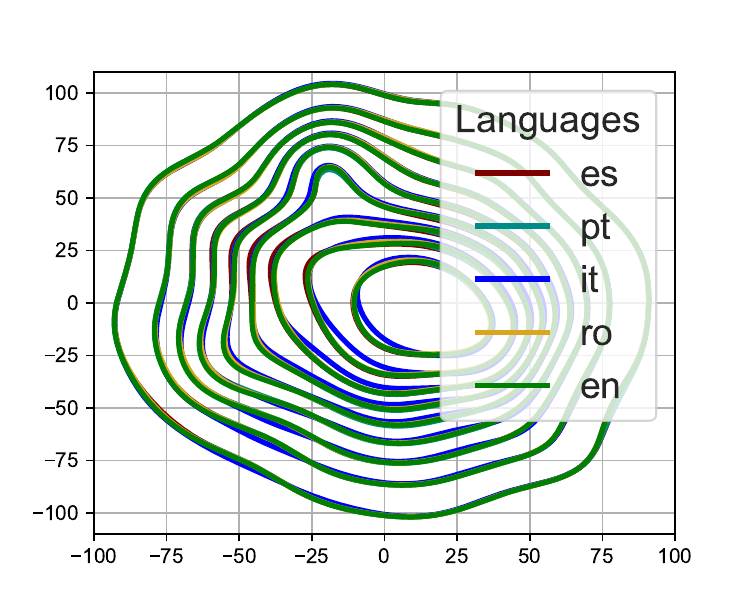}
        \caption{Semantic alignments}
        \label{fig:align}
      \end{subfigure}
    \caption{Different analytical methods lead to different conclusions.
    \ref{fig:cluster} means the target language family clusters the representations of translations from English (\texttt{en}) to other languages through the encoder.
    \ref{fig:align} indicates the encoder semantically aligns different source languages.
    Language codes in this work follow ISO 639-1, and Appendix \ref{appendix:introduction} provides details of those figures.}
    \label{fig:comparison}
\end{figure}

Representational analysis in MNMT models can guide the improvement of zero-shot translation.
However, two contrary opinions are demonstrated by prior works: (1) the encoder clusters translation representations based on the target language \cite{investigateMNMT,DisentPos-2021,zero-2023,transfer-2023,lcs-2024}, as illustrated in Figure \ref{fig:cluster}; (2) an ideal encoder is expected to learn language-agnostic representations, capturing general cross-lingual features that are transferable across languages \cite{Constras-2021,Regular-2022,Regular-2023,aganostic-2024}, as shown in Figure \ref{fig:align}.
In this work, we aim to analyze and reconcile this discrepancy.
We first introduce the identity pair, a pseudo pair translating a sentence to itself.
Specifically, the analyses conducted by those prior works rely on real translation pairs, leading to inaccurate results, as a translation pair cannot serve as a base measure for another pair.
The identity pair, however, addresses this issue by serving as a proxy for the optimal representation of a language instead of a translation pair.
Then, multiple analytical methods are employed to show the representation transfer within MNMT models.
Our findings offer a unified perspective on these two opinions: the encoder transfers translation representations into the target language subspace, where different source languages are semantically aligned.
Thus, the zero-shot translation deficiency stems from the failure to transfer the translation representation to the target language, as it becomes entangled with representations of other languages in the encoder.

Guided by our findings, we propose two methods for the encoder and decoder, respectively, to improve multilingual representations:
\textbf{Lo}w-Rank \textbf{L}anguage-specific \textbf{E}mbedding (\textsc{LoLE}) is applied to bias the representations in the subspaces of target languages at the encoder; and \textbf{L}anguage-specific \textbf{C}ontrastive \textbf{L}earning of \textbf{R}epresentations (\textsc{LCLR}) is applied at the decoder to isolate representational space across languages.
We evaluated the proposed methods on three benchmarks, Europarl-15 \cite{europarl}, TED-19 \cite{ted}, and OPUS-100 \cite{massive-2020, TLP-2021}, for two automatic metrics, SacreBLEU \cite{sacrebleu} and BERTScore \cite{bertscore}.
The experimental results show that our methods outperform strong baselines in training from scratch because of improved representational transferability.
Our methods also perform effectively in fine-tuning, even though pre-trained models are trained by different strategies of language tags, which proves that target language information on the encoder side consistently benefits MNMT.

\section{Background}\label{section:background}
\subsection{Multilingual Neural Machine Translation}\label{section:mnmt}
\citet{GooglesMNMT, TagMatter_2021} demonstrated that the training strategy of adding a language tag at the beginning of the input sentence on the encoder side boosts the zero-shot translation capacity of the MNMT model.
Given a multilingual corpus $\mathbb{C}$ that covers a set of $t$ languages, a set of their corresponding language tags exists:
$\mathbb{L}=\{l_1, l_2, \ldots, l_t\}$.
For a source-target sentence pair $(\boldsymbol{x}, \boldsymbol{y})$, i.e., 
$\boldsymbol{x}=x_1, x_2, \ldots, x_n$
and
$\boldsymbol{y}=y_1, y_2, \ldots, y_m$,
the training data consists of a pair in form of $(\boldsymbol{x}, l,\boldsymbol{y})$, where $l$ is the language tag of $\boldsymbol{y}$ that instructs translation from $\boldsymbol{x}$ into language $l$. The model is trained over all pairs in $\mathbb{C}$ to optimize the following cross-entropy loss:
\begin{equation}\label{eq1}
\mathcal{L}_{ce}= -\sum_{\boldsymbol{x}, l, \boldsymbol{y} \in \mathbb{C}} \log p(\boldsymbol{y} | l, \boldsymbol{x}; \theta),
\end{equation}
where $p(\boldsymbol{y} | l, \boldsymbol{x}; \theta)$ is the probability distribution of $\boldsymbol{y}$ and $\theta$ represents the model parameters.

\subsection{The Discrepancy in Prior Works}\label{section:priorworks}
\citet{Constras-2021, Regular-2023, aganostic-2024} state that, for an encoder-decoder MNMT model, an ideal encoder is regarded as transferring the source sentence into a language-agnostic state, preserving only semantic information.\footnote{Although \citet{Constras-2021} proposed that the ideal output of the encoder is language-agnostic by adding a source language tag at the beginning of the encoder, the follow-up works \cite{Regular-2023, aganostic-2024} practiced this concept with adding a target language tag, which is the main strategy investigated in this work.}
As evidence, the t-distributed stochastic neighbor embedding (t-SNE) \cite{tsne}, which can convert similarities between vectors into joint probabilities, has been used to show that representations of sentences from different languages are aligned at the output of the encoder when sharing the same semantics.
However, this result contrasts with the findings of \citet{investigateMNMT, transfer-2023, zero-2023}.
Specifically, using the singular value canonical correlation analysis (SVCCA) \cite{svcca} to compare the similarity between two vectors, i.e., the sentence representations of two translations, reveals that the encoder tends to transfer the representation into a state with target language features. 

We argue that this discrepancy stems from the lack of a base measure.
Namely, those works always compare the representations of real translation pairs in which different analysis methods lead to different results.
For instance, the translation from English to German, denoted by \texttt{en}$\to$\texttt{de}, cannot be accurately measured by comparing it with another translation from a different language \texttt{x}$\to$\texttt{de}, because \texttt{en}$\to$\texttt{de} is expected to be measured by the language representation of either \texttt{de} or \texttt{en}.
Thus, proposing a base measure is necessary to draw the same conclusion from different analysis methods, e.g., t-SNE or SVCCA\footnote{We follow \citet{DisentPos-2021} to measure SVCCA scores at the sentence level, which is introduced in Appendix \ref{appendix:svcca}.}.

\begin{figure*}[t]
    \centering
    \begin{subfigure}[b]{0.245\linewidth}
        \centering
        \includegraphics[width=\linewidth]{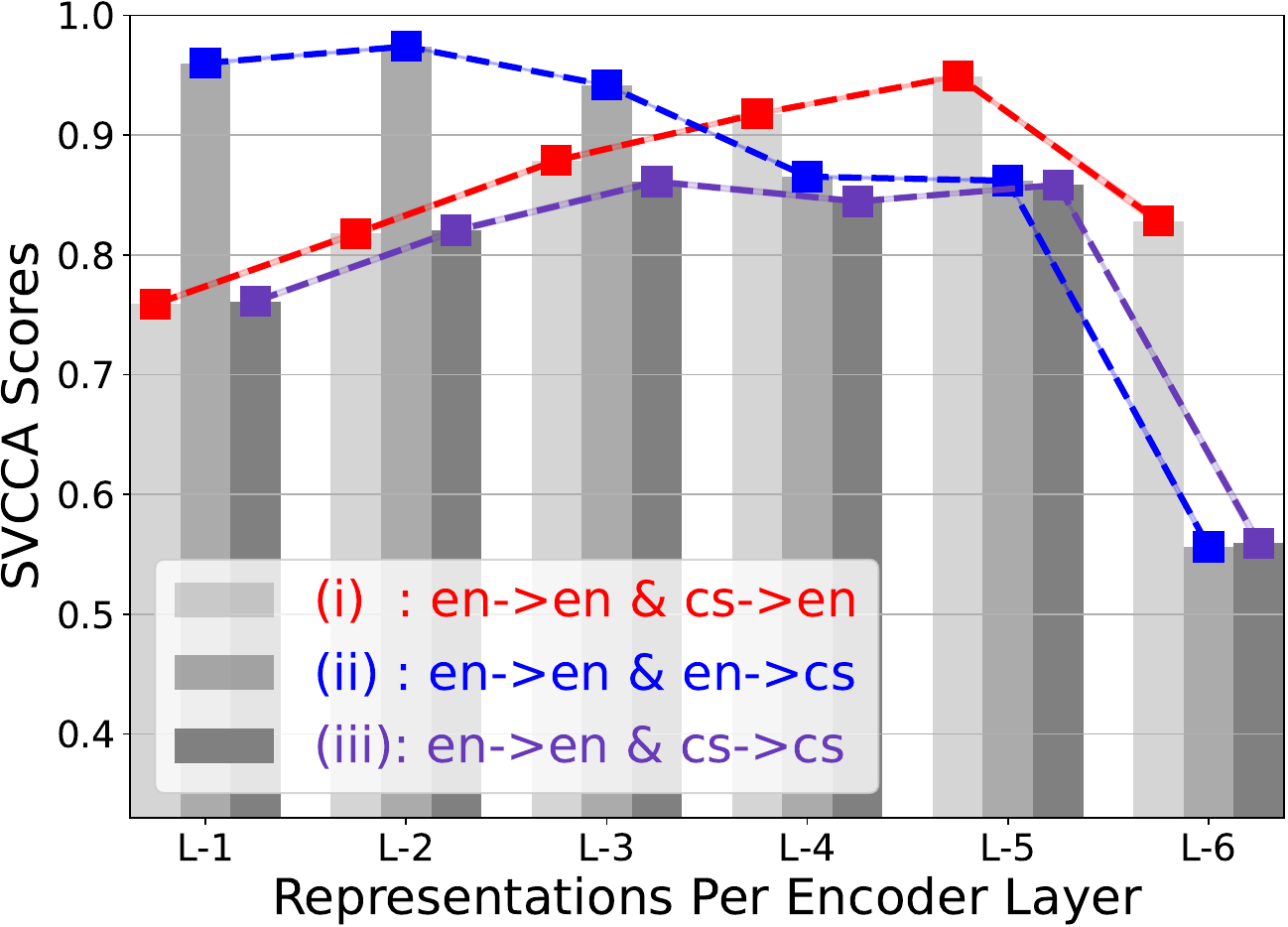}
        \caption{(\texttt{en}, \texttt{cs}) in Europarl.}
        \label{fig:transfer_1}
    \end{subfigure}
    \begin{subfigure}[b]{0.245\linewidth}
        \centering
        \includegraphics[width=\linewidth]{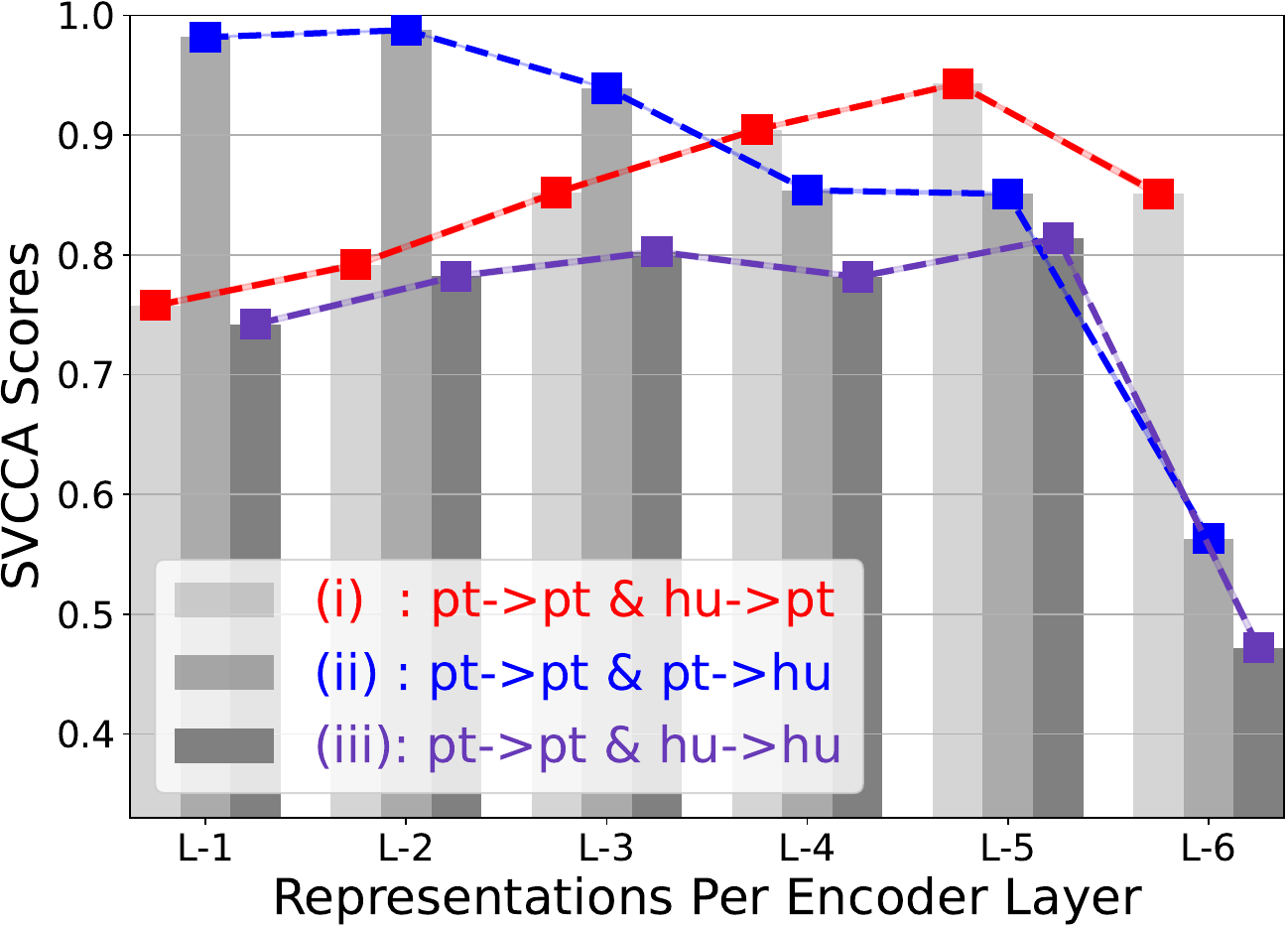}
        \caption{(\texttt{pt}, \texttt{hu}) in Europarl.}
        \label{fig:transfer_2}
    \end{subfigure}
    \begin{subfigure}[b]{0.245\linewidth}
        \centering
        \includegraphics[width=\linewidth]{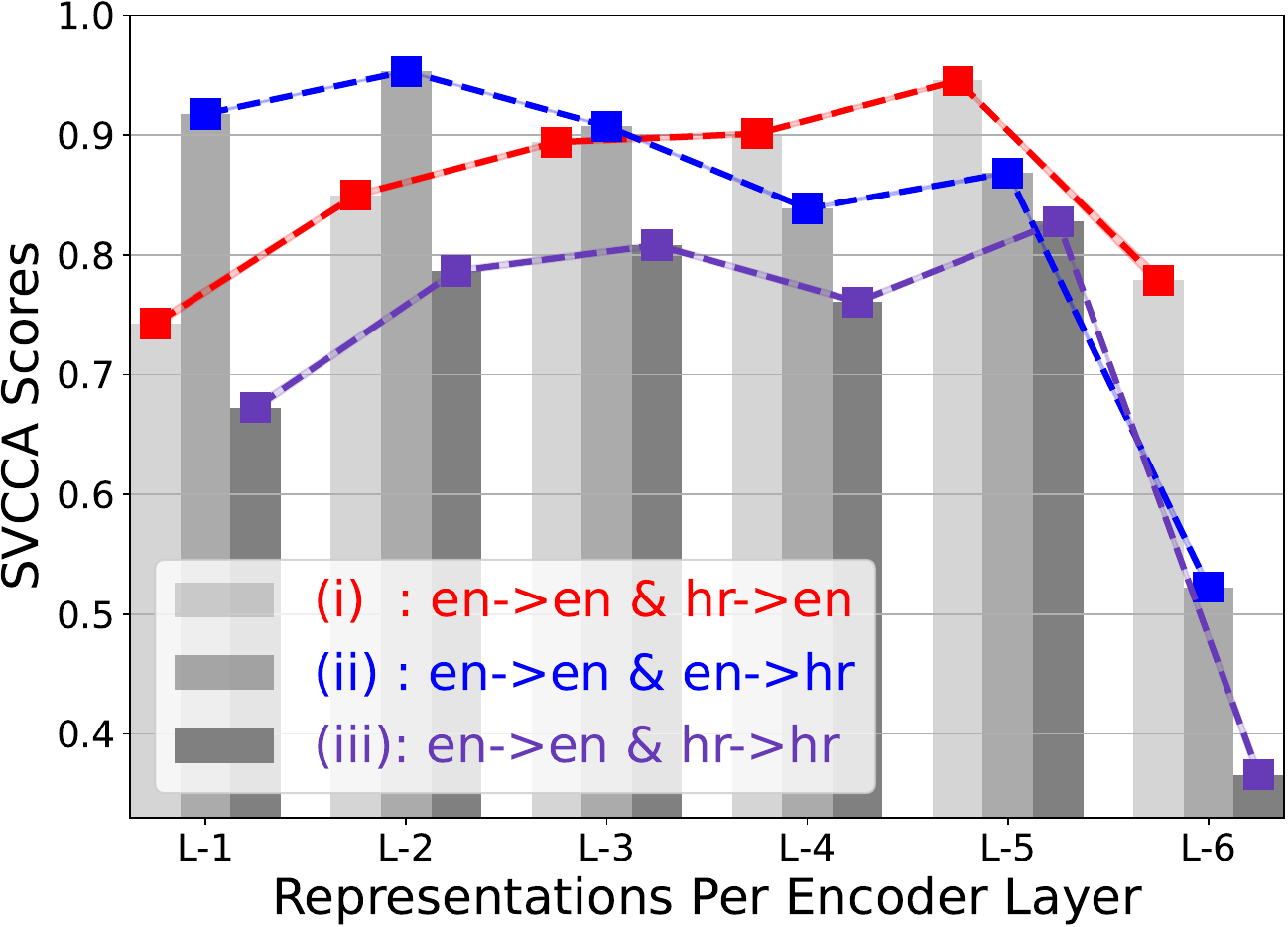}
        \caption{(\texttt{en}, \texttt{hr}) in TED.}
        \label{fig:transfer_3}
    \end{subfigure}
    \begin{subfigure}[b]{0.245\linewidth}
        \centering
        \includegraphics[width=\linewidth]{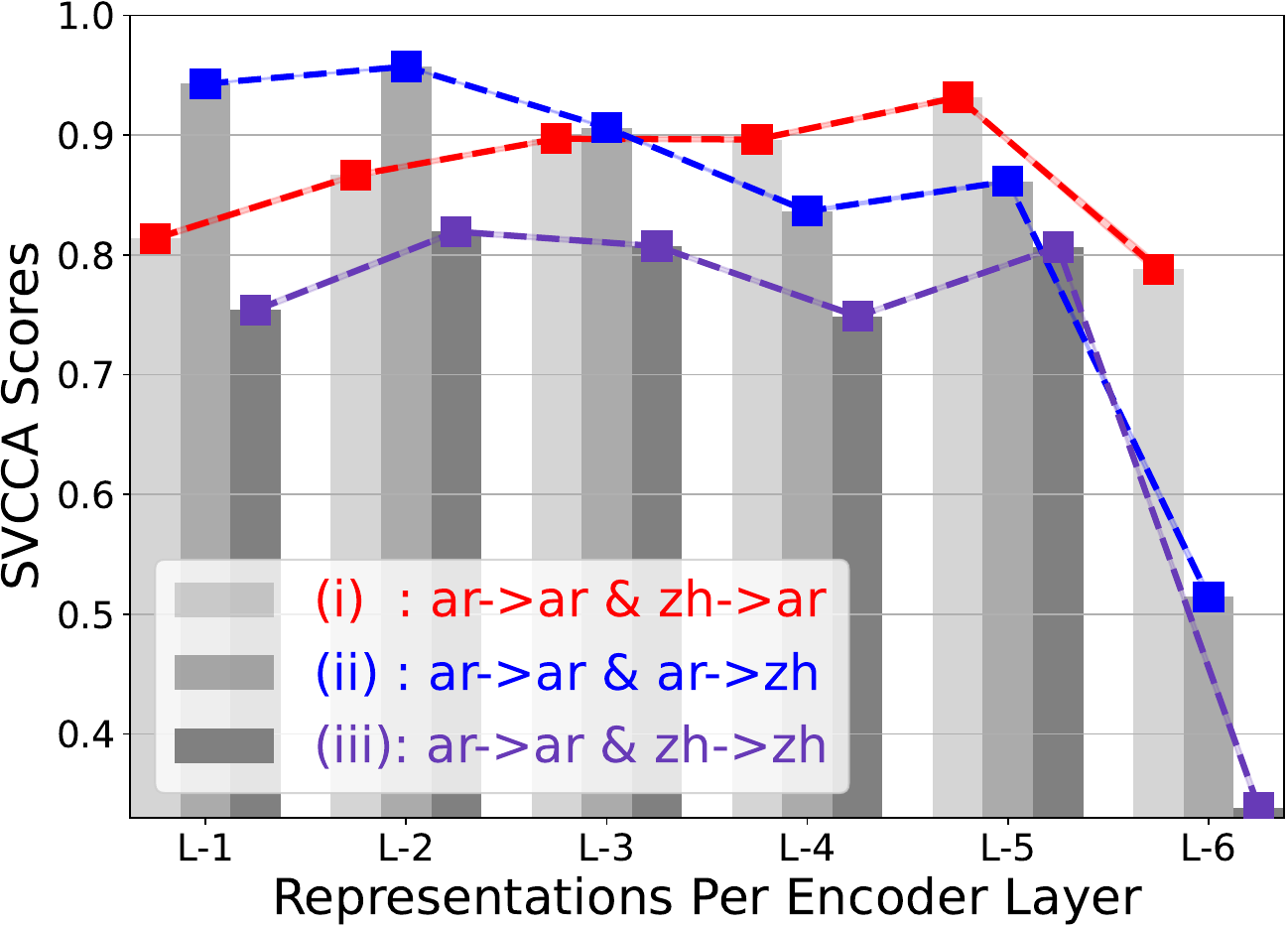}
        \caption{(\texttt{ar}, \texttt{zh}) in TED.}
        \label{fig:transfer_4}
    \end{subfigure}
    \caption{Visualizations of layer-wise SVCCA scores for the encoder.
    (\ding{172}, \ding{173}) indicate the source language and target language, respectively.
    The analyzed models have 6 encoder layers, and the analysis based on models with 8 and 10 encoder layers is shown in Figure \ref{fig:supplement}.}
    \vspace{-1em}
    \label{fig:transfer}
\end{figure*}
\begin{figure*}[t]
\centering
    \begin{subfigure}[b]{0.245\linewidth}
        \centering
        \includegraphics[width=\linewidth]{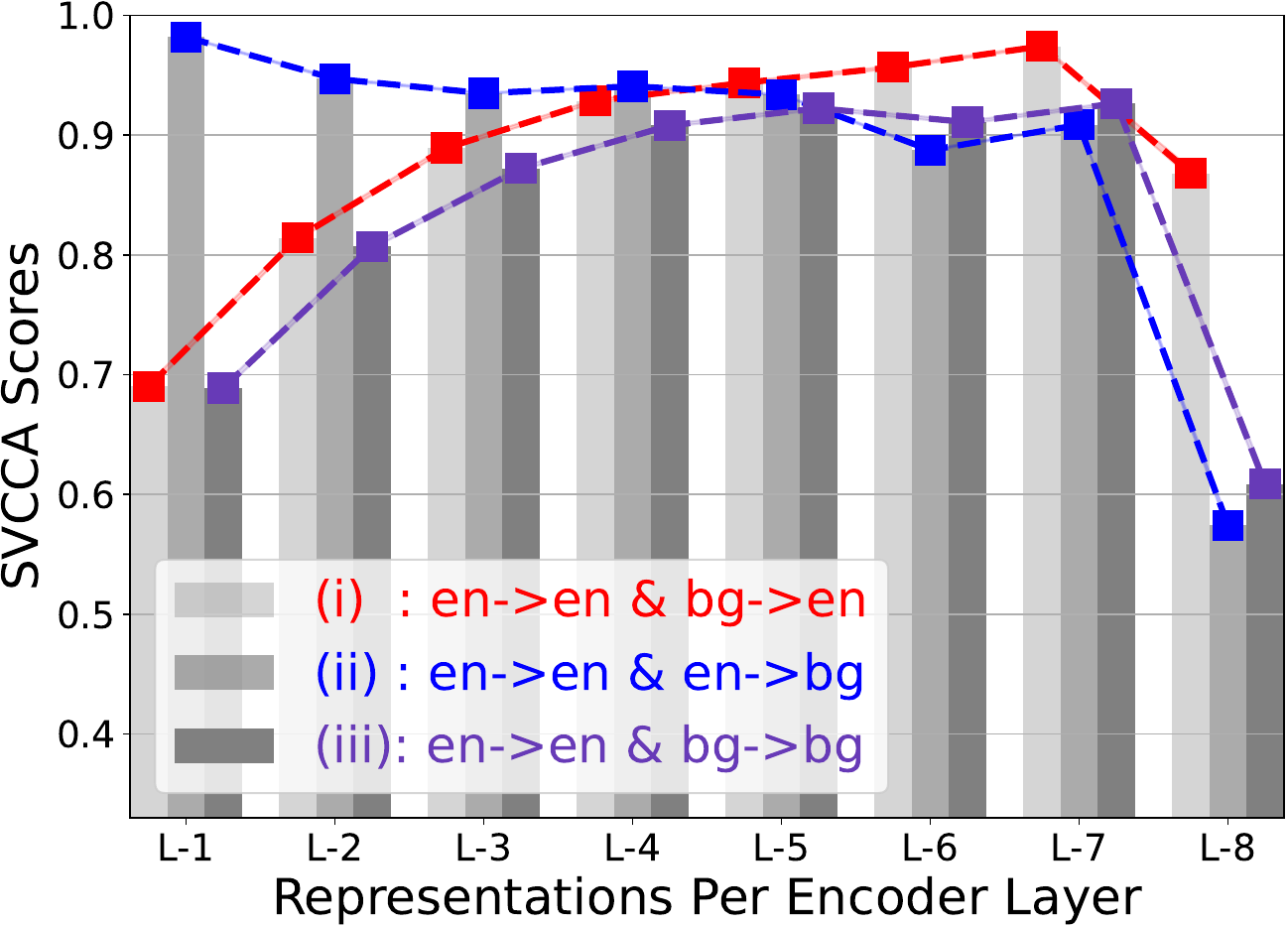}
        \caption{(\texttt{en}, \texttt{bg}) with 8 layers.}
        \label{fig:supplement_1}
    \end{subfigure}
    \begin{subfigure}[b]{0.245\linewidth}
        \centering
        \includegraphics[width=\linewidth]{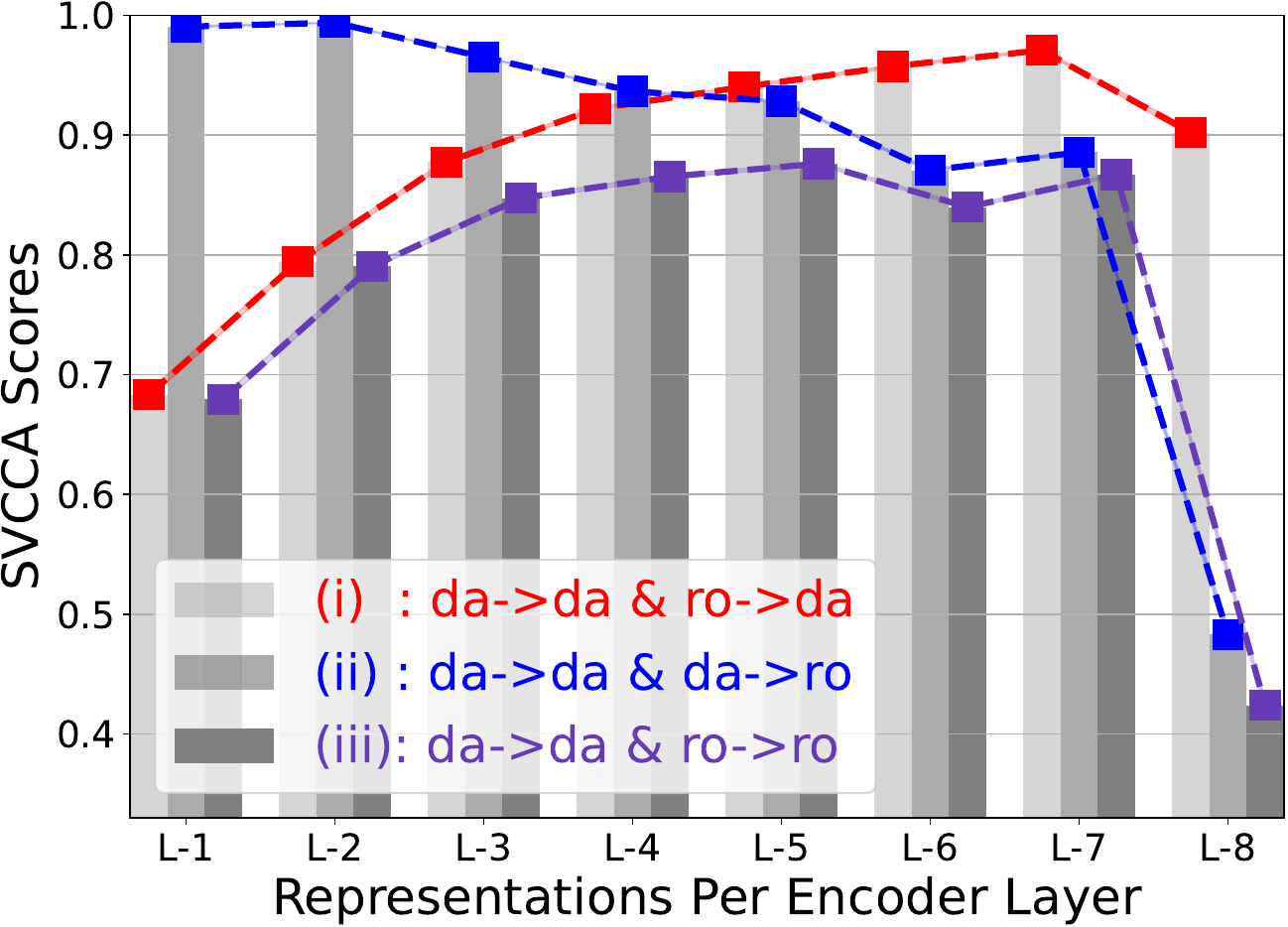}
        \caption{(\texttt{da}, \texttt{ro}) with 8 layers.}
        \label{fig:supplement_2}
    \end{subfigure}
    \begin{subfigure}[b]{0.245\linewidth}
        \centering
        \includegraphics[width=\linewidth]{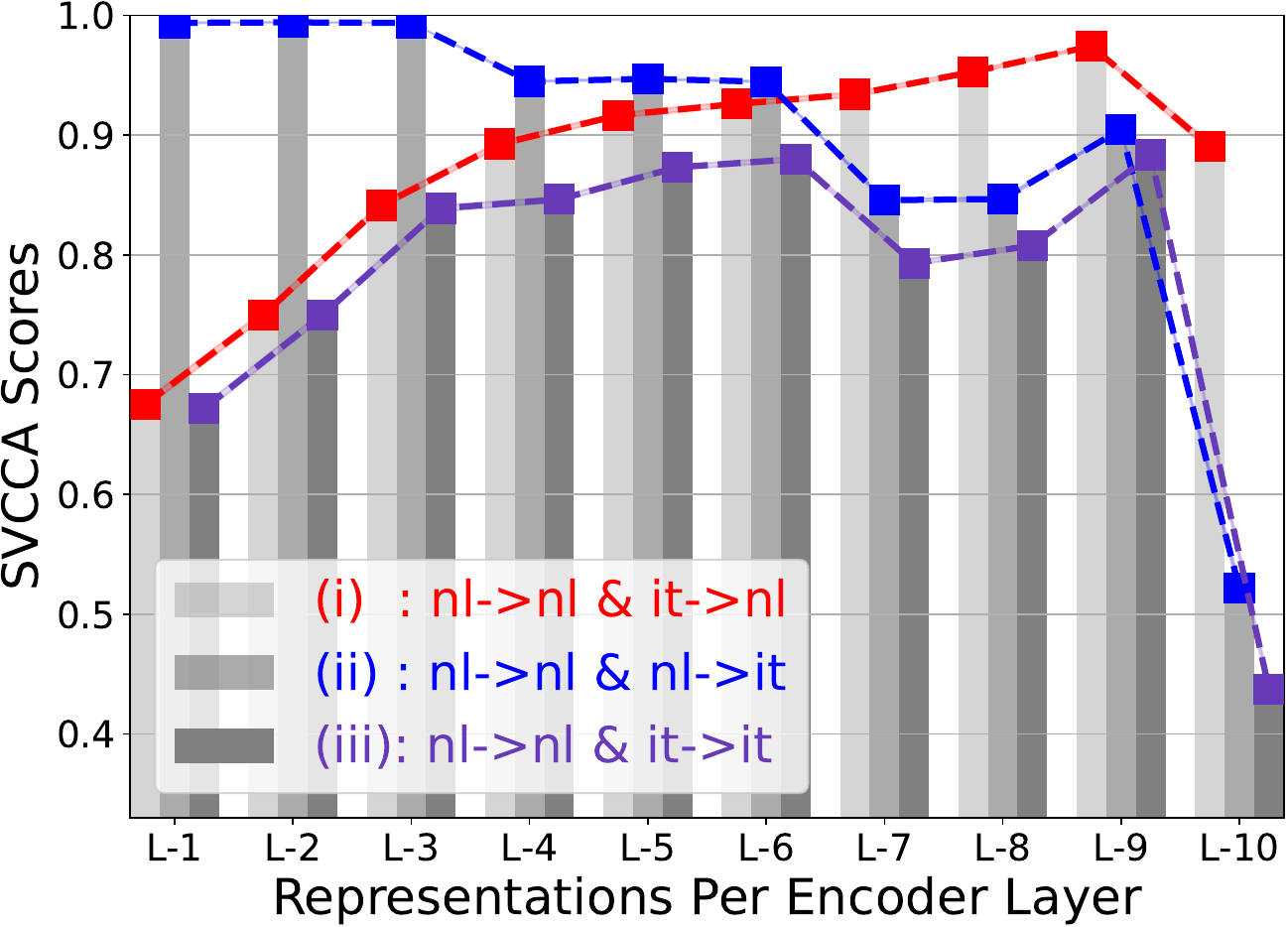}
        \caption{(\texttt{nl}, \texttt{it}) with 10 layers.}
        \label{fig:supplement_3}
    \end{subfigure}
    \begin{subfigure}[b]{0.245\linewidth}
        \centering
        \includegraphics[width=\linewidth]{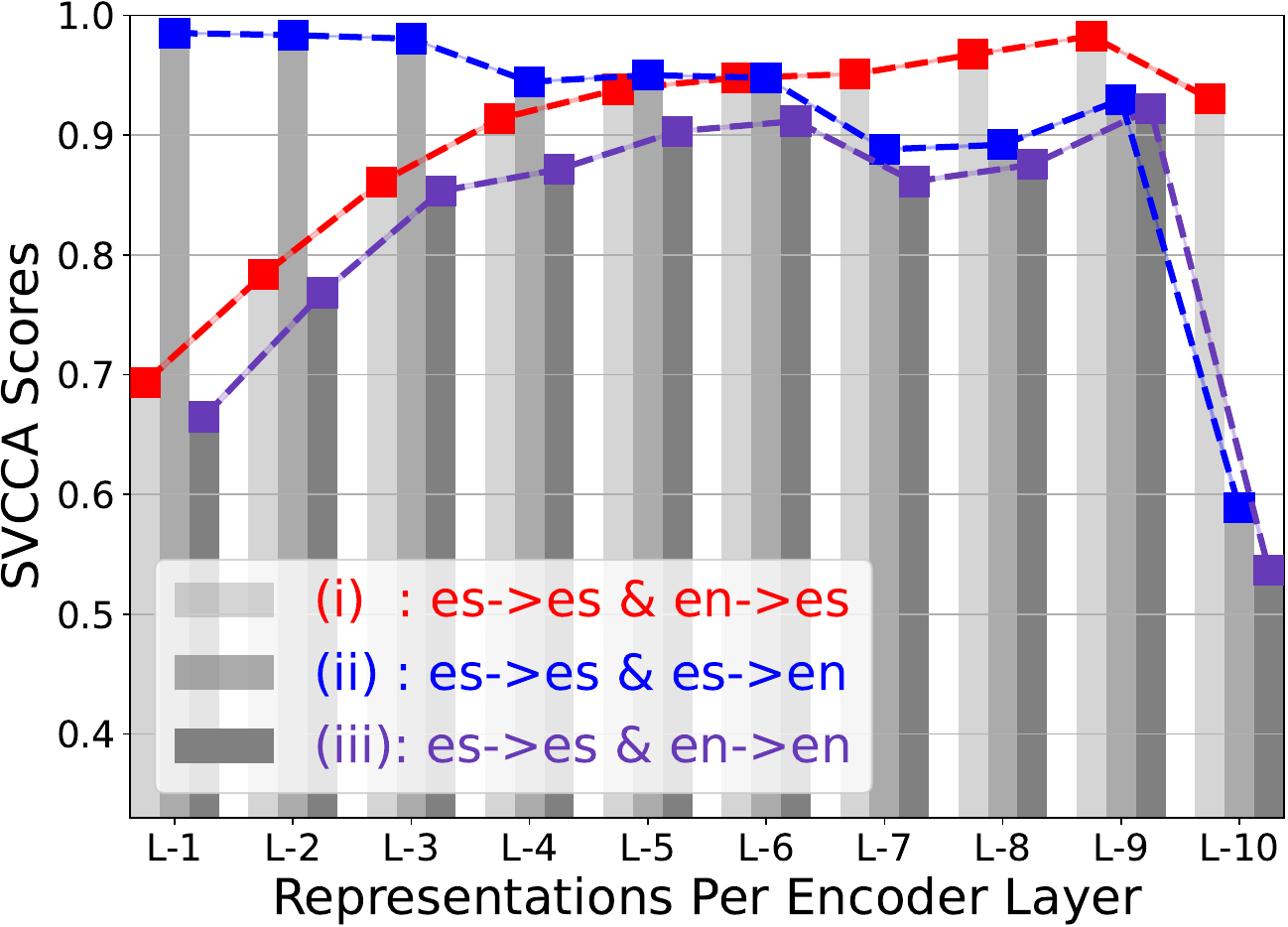}
        \caption{(\texttt{es}, \texttt{en}) with 10 layers.}
        \label{fig:supplement_4}
    \end{subfigure}
    \vspace{-1em}
\caption{Visualizations of layer-wise SVCCA scores for the encoders with 8 and 10 layers on diverse languages in Europarl-15, as a comparison of Figure \ref{fig:transfer} to prove the generalization.}
\label{fig:supplement}
\end{figure*}

\section{Investigating Representation Transfer in MNMT}\label{section:investigate}

We conduct preliminary experiments to investigate representations by introducing identity pairs as base measures using two different datasets, Europarl-15 \cite{europarl} and TED-19 \cite{ted}, which are introduced in Appendix~\ref{appendix:datasets} in detail.
Then, following \citet{investigateMNMT}, our investigation is based on Transformer models with 6 encoder and decoder layers.
We also investigate scenarios with 8 and 10 encoder layers.
Appendix~\ref{appendix:models} introduces the detailed model settings.

\subsection{Identity Pairs}
An identity pair refers to a sentence pair translating from one sentence into itself to represent the optimal state of processing language features, i.e., the semantics and syntax of the source sentence by the model.
Notably, our models are only trained by translating from one language to another.
In this setup, the identity pair is a zero-shot translation, which does not simply copy the input to the output.\footnote{This claim is supported by \citet{adaptingZero}, which demonstrate another zero-shot scenario: removing the language tag during inference results in any source sentence being translated into English. Thus, the identity pair indeed presents a translation process by adding a language tag.}
On the encoder side, we derive the representation from a language translating to itself, i.e., $(\boldsymbol{x}, l',\boldsymbol{x})$, where $l'$ is the language tag of $\boldsymbol{x}$, with the aim of recovering the source sentence from the hidden representations without inference on the decoder side.
We also derive the representation in the decoder from the gold translation of $(\boldsymbol{x}, l',\boldsymbol{x})$.

We use SacreBLEU \cite{bleu, sacrebleu} to evaluate the translation quality of 6 identity pairs, which are generated by inference.
The scores of \texttt{en}$\to$\texttt{en}, \texttt{de}$\to$\texttt{de}, and \texttt{pt}$\to$\texttt{pt} in Europarl-15 are 73.49, 61.04, and 71.97, which significantly outperform 44.04 of \texttt{de}$\to$\texttt{en}, 36.63 of \texttt{en}$\to$\texttt{de}, and 46.24 of \texttt{en}$\to$\texttt{pt}, respectively.
Similarly, \texttt{en}$\to$\texttt{en}, \texttt{tr}$\to$\texttt{tr}, and \texttt{vi}$\to$\texttt{vi} in TED-19 obtain scores of 72.52, 36.58 and 59.26, which are higher than 34.92 of \texttt{de}$\to$\texttt{en}, 14.81 of \texttt{en}$\to$\texttt{tr}, and 29.78 of \texttt{en}$\to$\texttt{vi}, respectively.
Such high scores in the identity pair are caused by that short sentences are recovered from hidden representations perfectly, and long sentences only have a few changes in word selection.
Such evidence suggests that identity pairs can serve as base measures for comparing representations because the identity pair is a proxy for the optimal representation of a language, specifically, \texttt{x}$\to$\texttt{en} are expected to be close to \texttt{en}$\to$\texttt{en} in the representational space.

\subsection{Language Transfer Within the Encoder}\label{section:transfer}
Given two languages \ding{172} and \ding{173}, we follow \citet{Constras-2021, DisentPos-2021} to obtain sentence-level representations for $\boldsymbol{x}$ in \ding{172} and $\boldsymbol{y}$ in \ding{173} by applying mean pooling over token representations.
We then organize the comparisons into three cases to analyze the variation in encoder representations:
(\romannumeral1) comparing $(\boldsymbol{x}, l^\text{\ding{172}}, \boldsymbol{x})$ and $(\boldsymbol{y}, l^\text{\ding{172}}, \boldsymbol{x})$ to show how target language features are encoded;
(\romannumeral2) comparing $(\boldsymbol{x}, l^\text{\ding{172}}, \boldsymbol{x})$ and $(\boldsymbol{x}, l^\text{\ding{173}}, \boldsymbol{y})$ to show how source language features are encoded; (\romannumeral3) comparing two different identities, $(\boldsymbol{x}, l^\text{\ding{172}}, \boldsymbol{x})$ and $(\boldsymbol{y}, l^\text{\ding{173}}, \boldsymbol{y})$\footnote{$(\boldsymbol{x}, l^\text{\ding{172}}, \boldsymbol{x})$ indicates the identity of \ding{172}, i.e., \ding{172}$\to$\ding{172}, and $(\boldsymbol{y}, l^\text{\ding{172}}, \boldsymbol{x})$ indicates a sentence of \ding{173} translating to the sentence of \ding{172} instructed by the language tag of \ding{172}, i.e., \ding{173}$\to$\ding{172}.}.

\begin{figure}[t]
    \centering
    \includegraphics[width=\linewidth]{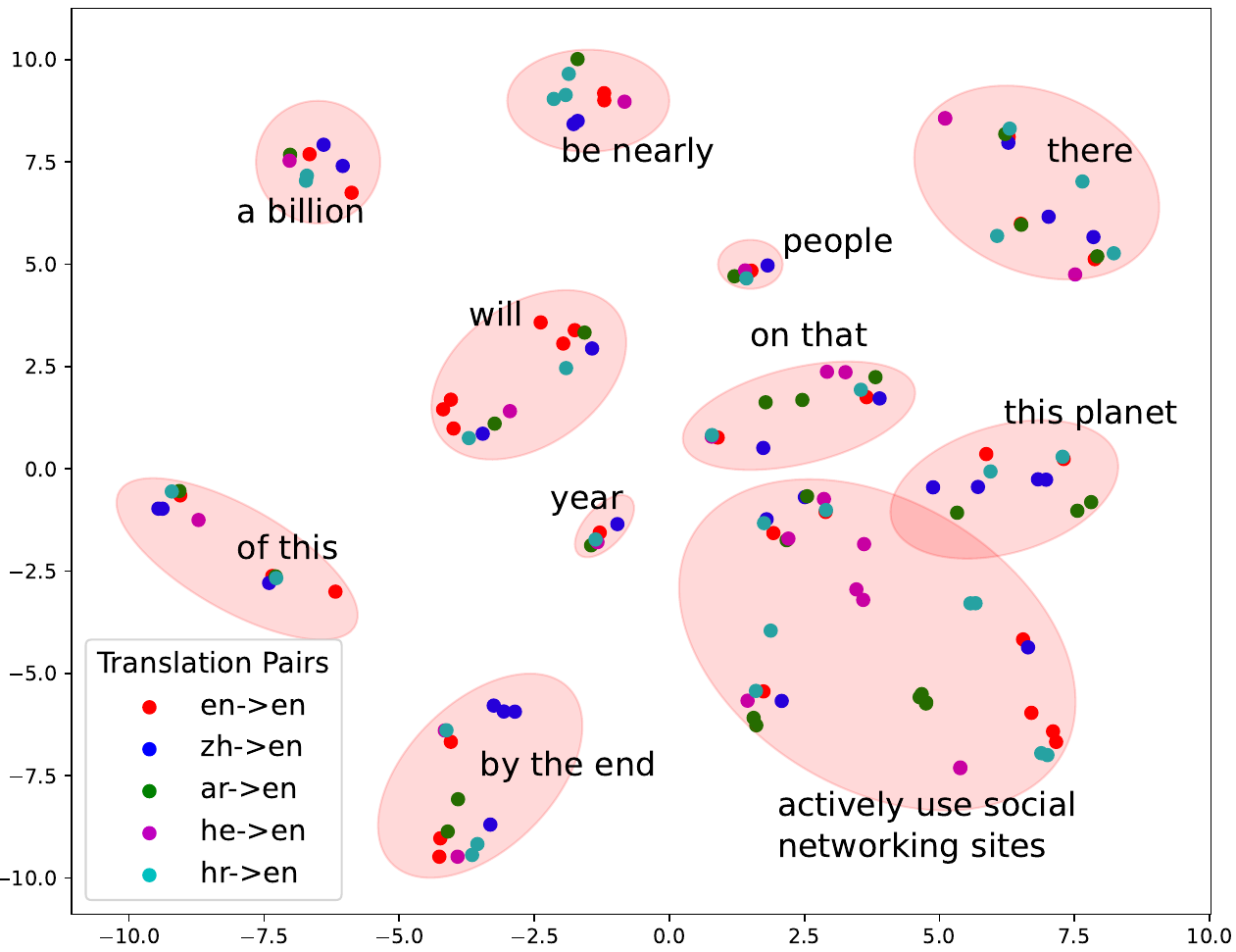}
    \caption{t-SNE plot of the token-level alignment between \texttt{en}$\to$\texttt{en} and \texttt{x}$\to$\texttt{en} in TED-19.
    Each point is a token's representation collected from the output of the encoder.
    Representations of different tokens are clustered by the semantics, which are denoted by English phrases, where the overall variance is 0.09.
    Appendix~\ref{appendix:alignment} shows the more details.}
    \label{fig:word_aligned}
\end{figure}

The two models trained by Europarl-15 and TED-19 show the same tendency in Figure \ref{fig:transfer}, i.e., the language features for \ding{172} of (\romannumeral1) consistently increase in both cases involving the central language, i.e., English, in Figures \ref{fig:transfer_1} and \ref{fig:transfer_3}, and non-central languages in Figures \ref{fig:transfer_2} and \ref{fig:transfer_4}. The target language feature of \ding{172} emerges as the primary factor that affects representations at the fifth and sixth layers when the cases of (\romannumeral1), (\romannumeral2), and (\romannumeral3) are compared.
Therefore, we can conclude that the language features of the representations are transferred to the target side within the encoder.
Meanwhile, we observe that the scores of (\romannumeral3) are close to or even exceed those of (\romannumeral2) at some layers both in Figure \ref{fig:transfer}.
This proves that the feature of the source language is not the primary factor for encoding representations because representations are transferred to the subspaces of target languages.
Thus, the comparison between (\romannumeral2) and (\romannumeral3) supports that language transfer is completed within the encoder.

To validate the generality of this conclusion, we extend our analysis to models with 8 and 10 encoder layers (Figure~\ref{fig:supplement}). The same trends hold: (\romannumeral1) continues to show increasing similarity scores, with the final values even higher than in the 6-layer setting, suggesting stronger target language alignment.
Again, the relationship between (\romannumeral2) and (\romannumeral3) remains consistent, further confirming that the source language is not the dominant factor in shaping encoder representations.
These results demonstrate that language transfer within the encoder is robust across different architectural depths.

On the other hand, identity pairs also allow the measurement of the alignment of different languages in the target language space through t-SNE.
Compared with the sentence-level measurement of \citet{Constras-2021,Regular-2023}, we measure the alignment of representations at the token level.
As shown in Figure \ref{fig:word_aligned}, semantic similarity causes the representations to cluster together.
Moreover, as shown in Appendix \ref{appendix:alignment}, these representations are not clustered before being processed by the encoder, and the case with different target languages has a higher overall variance.
Combined with the finding that the encoder transfers the representation of the source language to the target language, the evidence further suggests that there is no general and cross-lingual state for directly sharing semantic information within the encoder, and the alignment shown in Figure \ref{fig:align} occurs in the representational subspace of the target language.

\begin{figure}[t]
    \centering
    \begin{subfigure}[b]{0.32\linewidth}
        \centering
        \includegraphics[width=\linewidth]{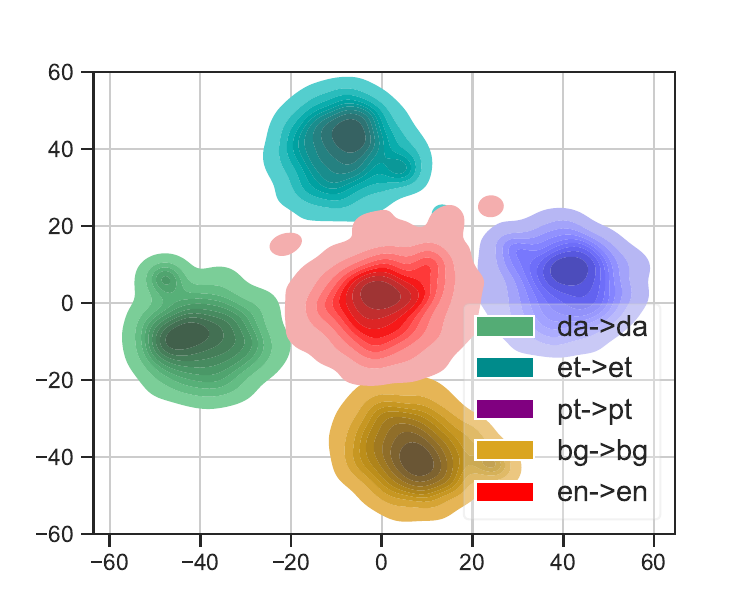}
        \caption{Identities.}
        \label{fig:entangle_1}
    \end{subfigure}
    \begin{subfigure}[b]{0.32\linewidth}
        \centering
        \includegraphics[width=\linewidth]{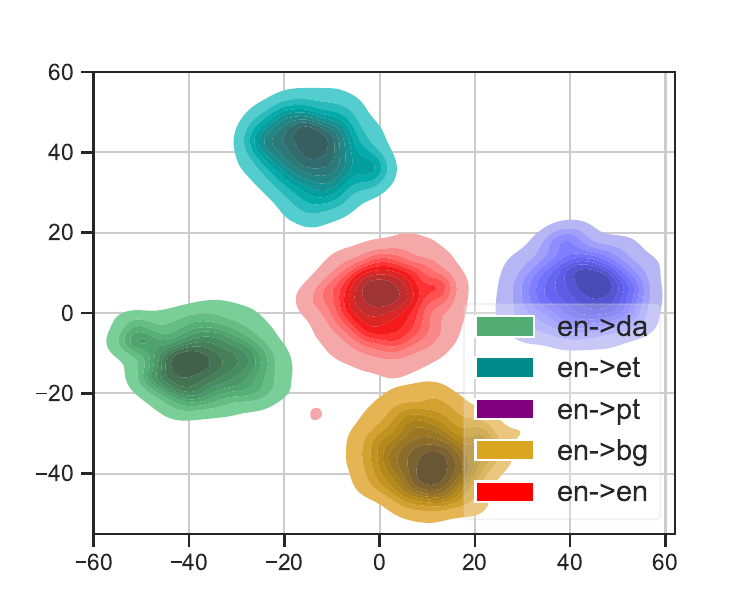}
        \caption{\texttt{en}$\to$\texttt{x}.}
        \label{fig:entangle_2}
    \end{subfigure}
    \begin{subfigure}[b]{0.32\linewidth}
        \centering
        \includegraphics[width=\linewidth]{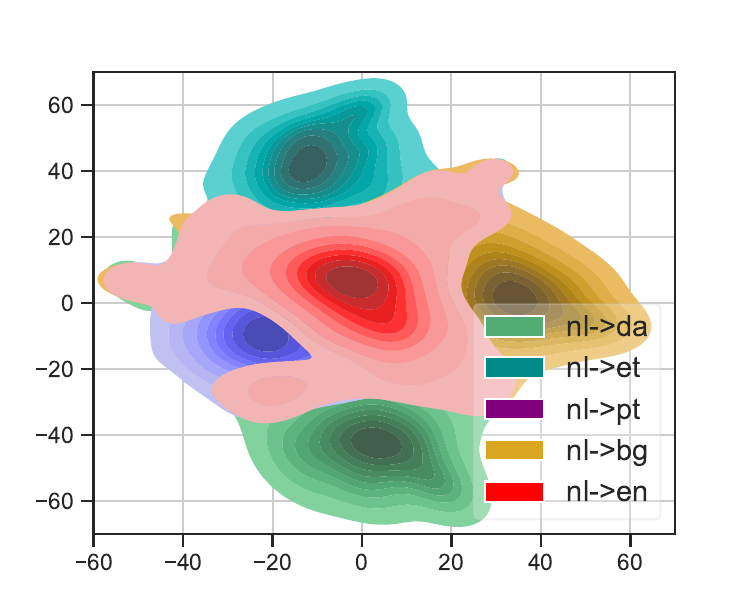}
        \caption{\texttt{nl}$\to$\texttt{x}.}
        \label{fig:entangle_5}
    \end{subfigure}
    \begin{subfigure}[b]{0.32\linewidth}
        \centering
        \includegraphics[width=\linewidth]{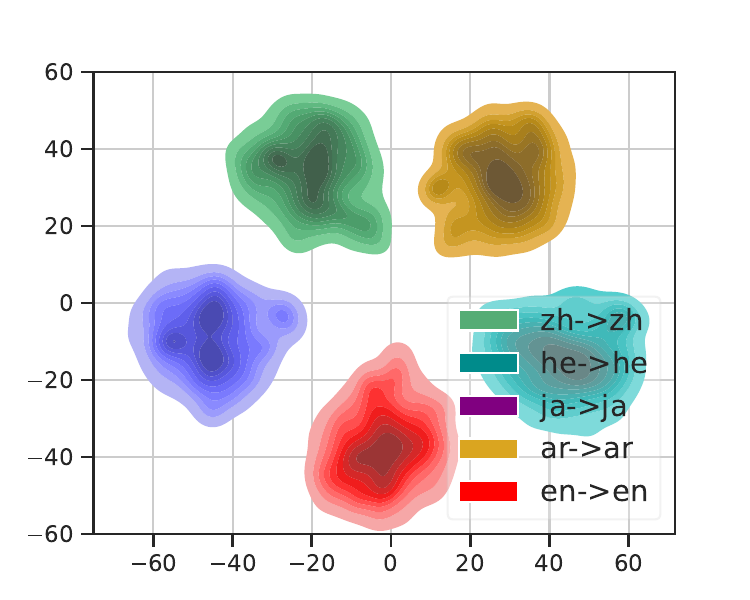}
        \caption{Identities.}
        \label{fig:entangle_3}
    \end{subfigure}
    \begin{subfigure}[b]{0.32\linewidth}
        \centering
        \includegraphics[width=\linewidth]{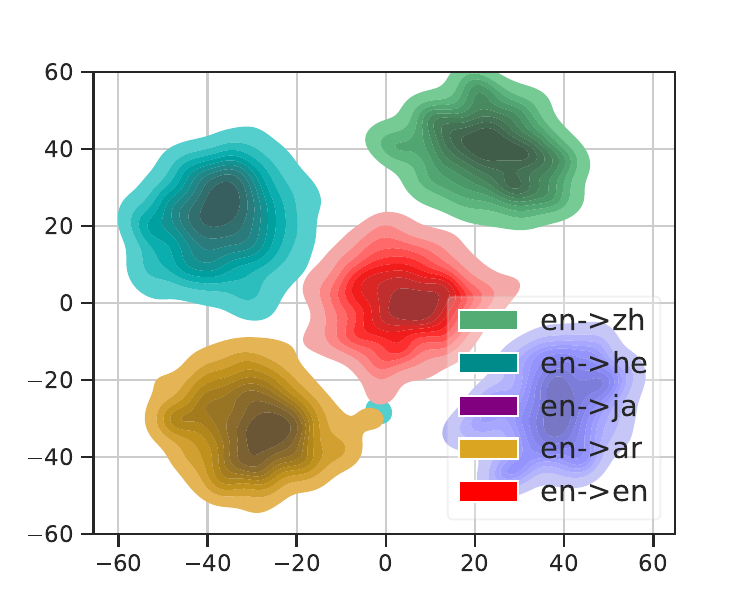}
        \caption{\texttt{en}$\to$\texttt{x}.}
        \label{fig:entangle_4}
    \end{subfigure}
    \begin{subfigure}[b]{0.32\linewidth}
        \centering
        \includegraphics[width=\linewidth]{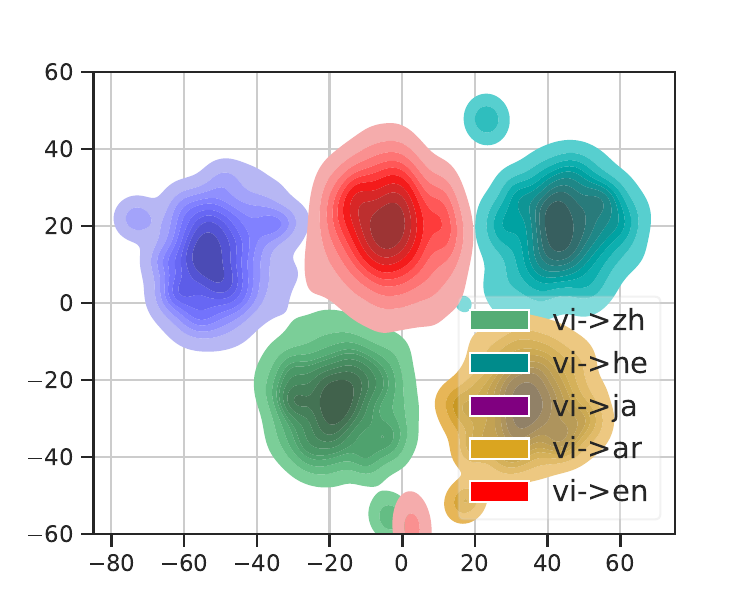}
        \caption{\texttt{vi}$\to$\texttt{x}.}
        \label{fig:entangle_6}
    \end{subfigure}
    \caption{Visualizations for the encoder's output by t-SNE and BiKDE. 
    \ref{fig:entangle_1}, \ref{fig:entangle_2} and \ref{fig:entangle_5} are measured in Europarl-15.
    \ref{fig:entangle_3}, \ref{fig:entangle_4} and \ref{fig:entangle_6} are measured in TED-19.
    }
    \label{fig:entangle}
\end{figure}
\subsection{Entanglements Hindering the Transfer}\label{section:entangle}
Although the investigation in Section \ref{section:transfer} shows that the representations gradually transfer to the target language in a translation pair, the representation spaces of multiple languages may potentially entangle with each other, resulting in the failure of the zero-shot translation \cite{adaptingZero}.
To further illustrate the relationship between different languages, we use t-SNE and BiKDE to visualize the representations at the output of the encoder for the several identity pairs in Europarl-15 and TED-19.
Figures \ref{fig:entangle_1} and \ref{fig:entangle_3} show that different identity pairs are uniformly distributed in the representational space.
This distribution proves again that the encoder is language-specific because each language has an isolated representational subspace.

Compared with identity pairs that represent the ideal capability of the model in processing languages, the distributions plotted in Figures \ref{fig:entangle_2} and \ref{fig:entangle_4} reflect the actual capacity for the supervised translation of \texttt{en}$\to$\texttt{x}.
Figures \ref{fig:entangle_2} and \ref{fig:entangle_4} show the distribution of representations in the pairs translating from \texttt{en}, which are similar to that of identity pairs.
The difference between identity pairs and supervised language pairs can be attributed to the influence of the source language information, which hinders the full use of the target language information learned by the encoder.

Moreover, the language-specific subspaces cannot be clearly separated for zero-shot translations, as shown in Figures \ref{fig:entangle_5} and \ref{fig:entangle_6}.
Specifically, all representations are entangled around the supervised language pair of \texttt{x}$\to$\texttt{en}, which hinders these representations from transferring into the ideal subspaces of the target language.
This aligns with \citet{adaptingZero} and \citet{transfer-2023} that multilingual representations are entangled, which explains the weakness of zero-shot translation compared with supervised translation, suggesting that improving the transferability of representations is attributed to the extent of language transfer within the encoder.

\begin{figure}[t]
    \centering
    \begin{subfigure}[b]{0.48\linewidth}
        \centering
        \includegraphics[width=\linewidth]{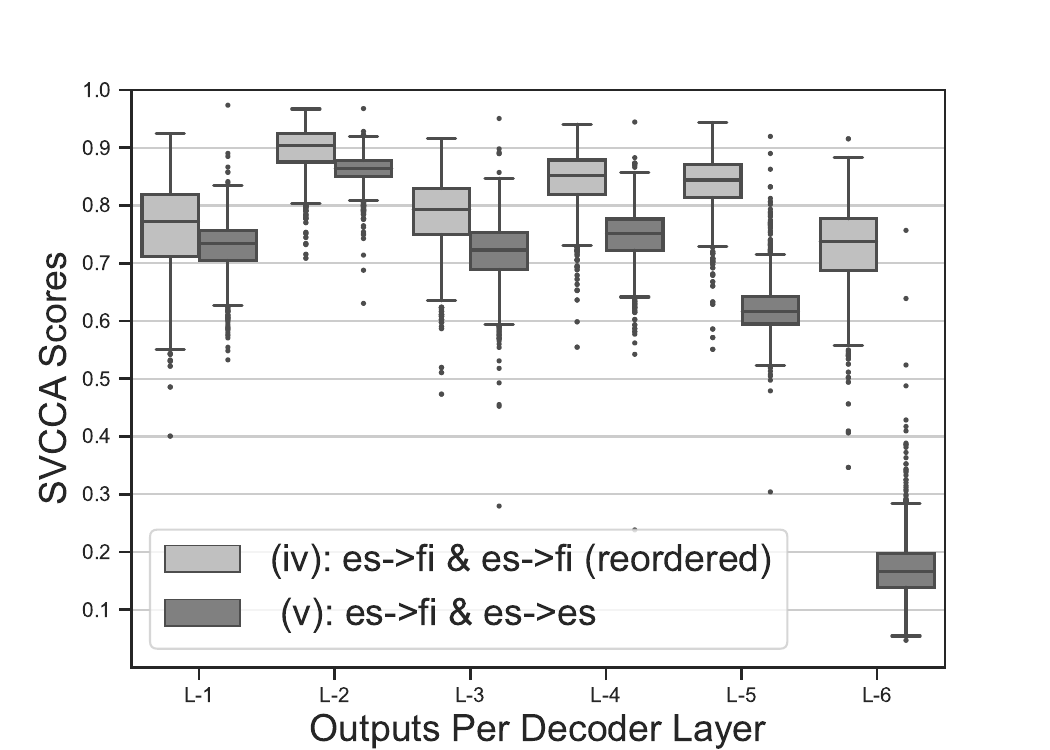}
        \caption{(\texttt{es}, \texttt{fi}) in Europarl.}
        \label{fig:decoder_1}
    \end{subfigure}
    \begin{subfigure}[b]{0.48\linewidth}
        \centering
        \includegraphics[width=\linewidth]{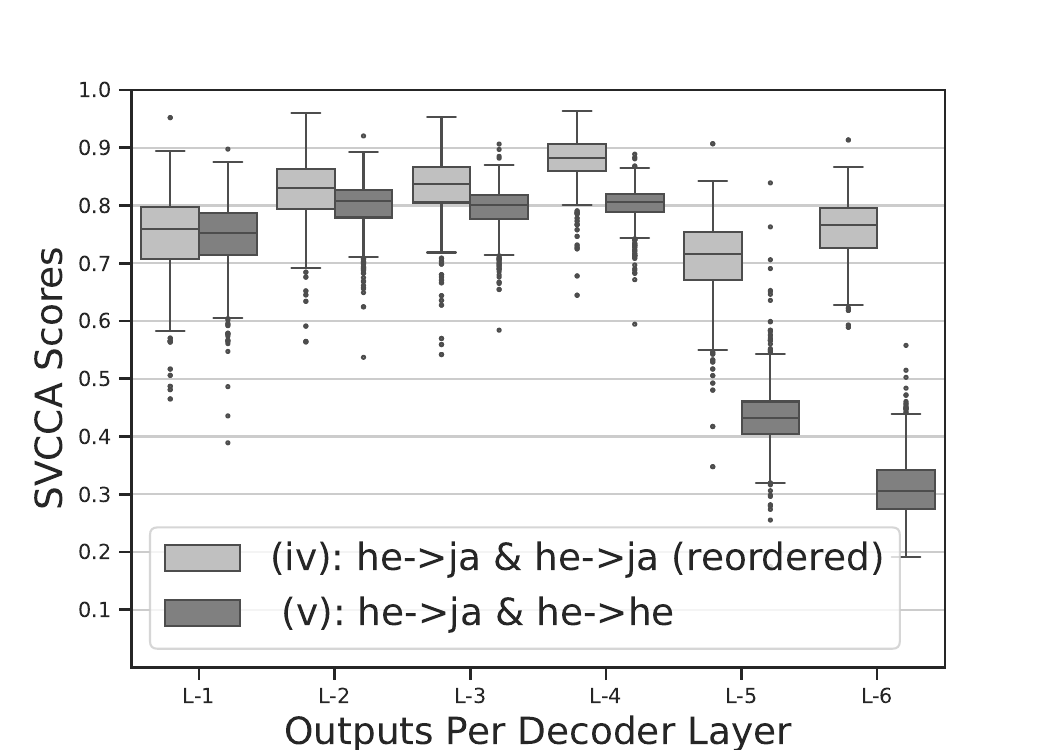}
        \caption{(\texttt{he}, \texttt{ja}) in TED.}
        \label{fig:decoder_2}
    \end{subfigure}
    \caption{Visualizations of layer-wise SVCCA scores for the decoder. (\ding{174}, \ding{175}) shows the involved languages.}
    \label{fig:decoder}
\end{figure}
\subsection{Language Features in the Decoder}\label{section:decoder}
We further investigate the importance of target language features versus semantics in the decoder.
Given two sentences $\boldsymbol{x}$ of language \ding{174} and $\boldsymbol{y}$ of language \ding{175}, the decoder representation of $(\boldsymbol{x}, l^\text{\ding{175}}, \boldsymbol{y})$ is considered as the base measure.
We group two cases:
(\romannumeral4)
For each sentence in a test set, we identify the pair $(\boldsymbol{x}', l^\text{\ding{175}}, \boldsymbol{y}')$ with the lowest SVCCA score in the encoder representation to derive a $\boldsymbol{x}'$ that is distant from $\boldsymbol{x}$.
Then, we compare it with the base measure to show the importance of target language features;
(\romannumeral5)
We compare the base measure and $(\boldsymbol{y}, l^\text{\ding{175}}, \boldsymbol{y})$ to show the importance of semantics.
The two scenarios shown in Figure \ref{fig:decoder} present the same trend, which is that (\romannumeral4) maintains high scores despite their semantics being entirely different.
At the top layers of the decoder, the gradually increasing difference between (\romannumeral4) and (\romannumeral5) confirms that the decoder tends to learn the target language specificity \cite{lsdecoder-2019}.
However, Figure \ref{fig:decoder} shows that, for (\romannumeral4), a wider interquartile range exists at the bottom layers of the decoder, and its scores are close to those of (\romannumeral5), which implies the weakness in distinguishing languages for zero-shot translations.

\section{Encouraging Representation Transfer}\label{section:4}
To validate the findings in Section \ref{section:investigate}, we propose two methods on the encoder and decoder sides, respectively, to improve transferability.
Based on the findings in Sections \ref{section:transfer} and \ref{section:entangle}, improving the extent of language transfer in the encoder can overcome the hindered representations of zero-shot language pairs.
We introduce a learnable embedding referred to as Low-rank Language-specific Embedding (\textsc{LoLE}).
It serves as biases to force representations to transfer into the target language with negligible cost.
Based on the findings in Section \ref{section:decoder}, the capacity for multilingual features is insufficient at the lower layers of the decoder.
We introduce Language-specific Contrastive Learning of Representations (\textsc{LCLR}) as an training extra task to regularize the representations to specify the representational boundary for each language.

\subsection{Low-Rank Embedding for the Encoder}\label{section:4.1}
\begin{figure}[t]
    \centering
        \begin{subfigure}[b]{0.45\linewidth}
        \centering
        \includegraphics[width=\linewidth]{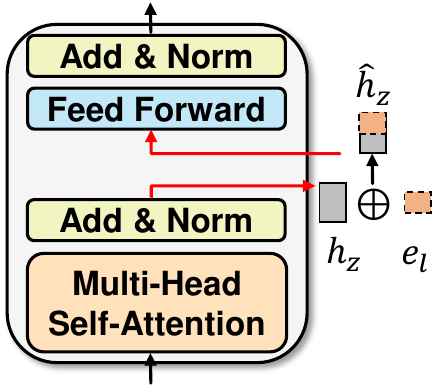}
        \caption{Process}
        \label{fig:method_1}
        \end{subfigure}
        \begin{subfigure}[b]{0.45\linewidth}
        \centering
        \includegraphics[width=\linewidth]{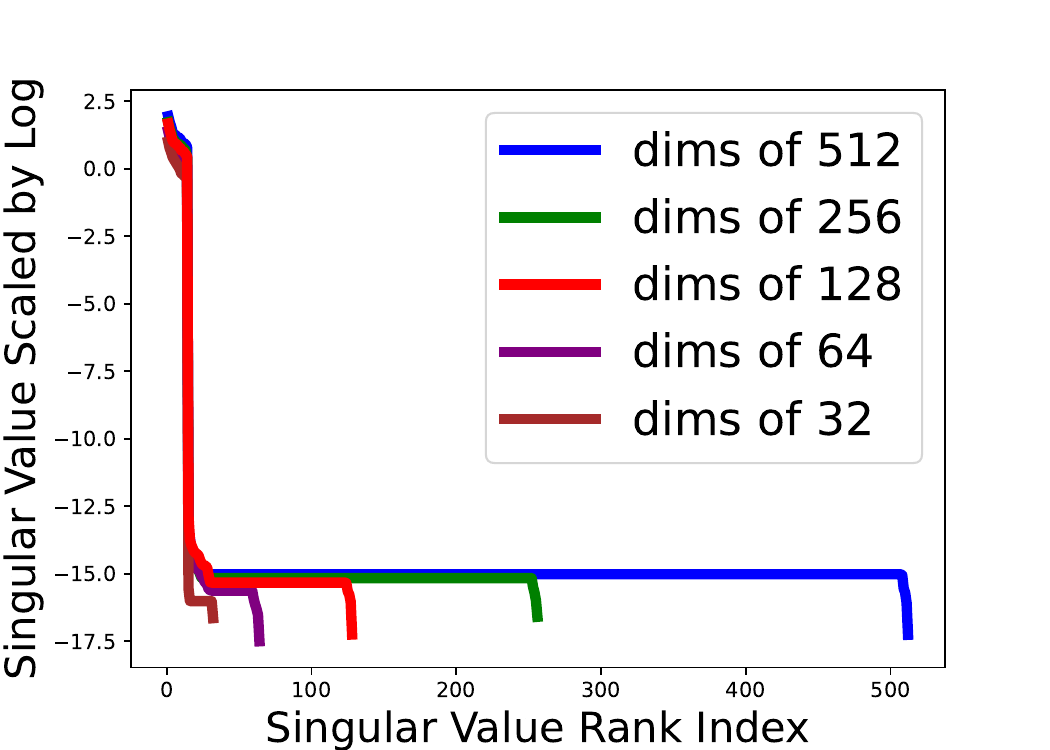}
        \caption{Dimension}
        \label{fig:method_2}
        \end{subfigure}
    \caption{Illustrations of LoLE. \ref{fig:method_1} shows the process in an encoder layer, where $\bigoplus$ represents the operation of Equation(\ref{eq3}) at a low rank. \ref{fig:method_2} is the spectrum with different dimensions of $\mathbb{E}$, which illustrates the singular values of the covariance matrix of $\mathbb{E}$ in sorted order and logarithmic scale.}
    \label{fig:method}
\end{figure}

Let $\mathbb{E} = \{ \boldsymbol{e}^1, \boldsymbol{e}^2, \ldots, \boldsymbol{e}^p \}$, $\boldsymbol{e}^j \in \mathbb{R}^{d}$, be a set of embeddings that correspond one-to-one with the languages in $\mathbb{L}$.
For a translation $(\boldsymbol{x}, l ,\boldsymbol{y})$, the embedding in $\mathbb{E}$ corresponding to $l$ is denoted by $\boldsymbol{e}^l$.
The hidden representation $\mathbf{H}^z =\{\boldsymbol{h}_1^z, \boldsymbol{h}_2^z, \dots,\boldsymbol{h}_q^z\}$, where $\mathbf{H}^z \in \mathbb{R}^{q \times d}$, is extracted before the feed-forward network (FFN) \cite{knn,knnStudy,knnSubset} at the $z$-th encoder layer.
Then, we broadcast $\boldsymbol{e}^l$ to $\mathbf{E}^l$, $\mathbf{E}^l \in \mathbb{R}^{q \times d}$, and we bias $\mathbf{H}^z$ to $\hat{\mathbf{H}}^z$:
\begin{equation}\label{eq3}
\hat{\boldsymbol{h}}^z_i = \boldsymbol{h}^z_i + \boldsymbol{e}^l_i,
\end{equation}
where $\hat{\mathbf{H}}^z$ is the input for the FFN of the $z$-th encoder layer (Figure \ref{fig:method_1}).
We execute this biasing at the second-top encoder layer to ensure sufficient capacity for fusing representations and language information, while implicitly allowing lower layers to focus on surface-level information.

The simple language categorization by embedding may lead to a risk of dimensional collapse in the latent space \cite{collapse}.
Thus, we reduce the dimension of $\mathbb{E}$ to $d^{e}$ to allow biasing in a low rank, and add it to the head of $\boldsymbol{h}_i^z$ to simultaneously encourage language transfer and minimize the influence on representations \cite{lora}.
Figure \ref{fig:method_2} is a spectrum used to illustrate dimensional collapse using a comparison of different $d^{e}$ in Europarl-15.
The spectrum shows that larger dimensions are primarily composed of noise, whereas a dimension that is too small adversely affects the learning of key features.
\begin{table*}[!ht]
  \centering
  \resizebox{\textwidth}{!}{
    \begin{tabular}{lcccccccccccccccc}
    \toprule
    & \multicolumn{6}{c}{Europarl-15} & \multicolumn{6}{c}{TED-19} & \multicolumn{4}{c}{OPUS-100} \\
    \midrule
    & \multicolumn{3}{c}{BLEU} & \multicolumn{3}{c}{B.S.} & \multicolumn{3}{c}{BLEU} & \multicolumn{3}{c}{B.S.} & \multicolumn{3}{c}{BLEU} & B.S. \\
    \midrule
    Method & \texttt{en}$\to$ & $\to$\texttt{en} & zero. & \texttt{en}$\to$ & $\to$\texttt{en} & zero. & \texttt{en}$\to$ & $\to$\texttt{en} & zero. & \texttt{en}$\to$ & $\to$\texttt{en} & zero. & \texttt{en}$\to$ & $\to$\texttt{en} & zero. & zero. \\
    \midrule
    \textsc{Vanilla} & 37.49 & 43.39 & 24.65 &88.50 & 95.71 & 84.27 & \textbf{24.53} & 29.67 & 11.98 & \textbf{83.77} & 93.54 & 77.74 & 23.37 & 28.30 & 5.04 & 69.98 \\ 
    \textsc{DisPos} & 37.15 & 43.37 & 25.89 & 88.39 & \textbf{95.72} & 84.69 & 24.08 & 29.43 & 12.80 & 83.62 & 93.49 & 78.36 & 22.72 & 28.24 & 5.58 & 70.74 \\
    \textsc{TLP} & 37.41 & 43.28 & 24.96 &88.47 & 95.71 & 84.40 & 24.44 & 29.62 & 12.74 & 83.73 & 93.53 & 78.24 & \textbf{23.41} & 28.30 & 4.60 & 69.40 \\
    \textsc{SemAli} & 37.27 & 43.06 & 25.25 & 88.42 & 95.69 & 84.43 & 23.55 & 28.67 & \hspace{0.4em}\textbf{13.45}$^\dag$ & 83.43 & 93.36 & \hspace{0.4em}\textbf{78.91}$^\dag$ & 22.35 & 28.29 & 6.42 & 72.00 \\
    \midrule
    \textsc{LoLE} &37.62& 43.50 & \hspace{0.4em}26.09$^\dag$ &\textbf{88.51} & \textbf{95.72} & 84.81 & 24.39 & 29.72 & 13.20 &83.74 & 93.54 & 78.65 & 23.15 & 28.28 & \hspace{0.4em}7.92$^\dag$ & \hspace{0.4em}\textbf{73.32}$^\dag$ \\
    \textsc{LCLR} & 37.44 & 43.43 & 25.71 &88.46 & \textbf{95.72} & 84.66 & 24.46 & 29.66 & 12.12 & 83.76 & 93.54 & 77.87 & 23.34 & \textbf{28.37} & 5.11 & 70.04 \\
    \textsc{BOTH} & \textbf{37.67} & \textbf{43.51} & \hspace{0.4em}\textbf{26.20}$^\dag$ & 88.50 & \textbf{95.72} & \hspace{0.4em}\textbf{84.85}$^\dag$ & 24.49 & \textbf{29.79} & \hspace{0.4em}13.31$^\dag$ & 83.76 & \textbf{93.56} & \hspace{0.4em}78.76$^\dag$ & 23.40 & 28.27 & \hspace{0.4em}\textbf{7.97}$^\dag$& \hspace{0.4em}73.10$^\dag$ \\
    \bottomrule
    \end{tabular}
  }
  \caption{Averaged scores for experiments of training from scratch.
           \textsc{BOTH} means using \textsc{LoLE} and \textsc{LCLR} together;
           \texttt{en}$\to$ and $\to$\texttt{en} abbreviates \texttt{en}$\to$\texttt{x} and \texttt{x}$\to$\texttt{en}; zero. means zero-shot language pairs; and B.S. abbreviates BERTScore.
           We only report zero-shot language pairs of OPUS-100 because BERTScore does not support some pairs in supervised translations, but zero-shot translation pairs of OPUS-100 are involved only with 6 languages, which are supported.
           The bold number indicates the best result and the numbers with $\dag$ are significantly better than \textsc{Vanilla} according to the significance test with $p <$ 0.05.
           The off-target ratios are reported in Appendix \ref{appendix:offratio}.
  }
  \label{tab:1}
\end{table*}

\subsection{Contrastive Learning for the Decoder}\label{section:4.2}
Given a training batch, we extract hidden representations from the output of each decoder layer and apply averaged pooling to obtain a fixed-dimensional representation for each sentence.
To avoid dimensional collapse \cite{understandContras,collapse}, we also use the head of the representation for contrastive learning, i.e., the vectors in the batch $\mathbb{B} = \{ \overline{\boldsymbol{h}}_1, \overline{\boldsymbol{h}}_2, \dots \}$, $\overline{\boldsymbol{h}}_i \in \mathbb{R}^{d^h}$, $d^h < d$.

To prevent a potential invalid training objective in sampling caused by the skewed distribution in a batch, we first define $\mathbb{B}' \subseteq \mathbb{B}$ by omitting instances that do not share their target language with any other instance in $\mathbb{B}$.
For a given instance of $\boldsymbol{h}^{\text{anc}} \in \mathbb{B}'$, which is the anchor in contrastive learning, we let $\mathbb{B}^+$ denote the subset of $\mathbb{B}'$, including instances with the same target language as $\boldsymbol{h}^{\text{anc}}$, where $ |\mathbb{B}^+|  > 1$.
Likewise, we define a subset for negative instance $\mathbb{B}^- = \mathbb{B}' \setminus \mathbb{B}^+$.
For contrastive learning, we randomly sample the positive instance $\boldsymbol{h}^{\text{pos}}$ from $\mathbb{B}^+$ and sample $k$ negative instances $\boldsymbol{h}^{\text{neg}}$ from $\mathbb{B}^-$.
Additionally, if $k > |\mathbb{B}^- | $, we dynamically clip $k$ to $ |\mathbb{B}^- |$.
Formally, the objective of \textsc{LCLR} is formulated as
\begin{equation}
\label{eq4}
\begin{split}
    \mathcal{L}_{ctr} &= -
    \sum_{\boldsymbol{h}^{\text{anc}} \in \mathbb{B}'}
    \log
    \frac
    {e^{s^+}}
    {e^{s^+} + \sum_{i=1}^k e^{s_i^-}}, \\
    s^+ &= \operatorname{sim}(\boldsymbol{h}^{\text{anc}}, \boldsymbol{h}^{\text{pos}}), \boldsymbol{h}^{\text{pos}} \in \mathbb{B}^+,
    \\
    s_i^- &= \operatorname{sim}(\boldsymbol{h}^{\text{anc}}, \boldsymbol{h}_i^{\text{neg}}), \boldsymbol{h}_i^{\text{neg}} \in \mathbb{B}^-,
\end{split}
\end{equation}
where $\operatorname{sim}(\cdot)$ calculates the similarity of representations using the cosine similarity.
The final training objective is the sum of Equations \ref{eq1} and \ref{eq3}, i.e.,
\begin{equation}
    \mathcal{L} = \mathcal{L}_{ce} + \mathcal{L}_{ctr}.
\end{equation}

\section{Experiments}\label{section:experiments}
\subsection{Setup} \label{section:setup}
\paragraph{Datasets}
Our experiments comprise three popular English-centric datasets, i.e., the training and validation sets only involving translation pairs translating to \texttt{en} or from \texttt{en}, including Europarl-15 \cite{europarl, mmcr4nlp}, TED-19 \cite{ted} and OPUS-100 \cite{massive-2020,TLP-2021}.
The details of those datasets can be found in Appendix \ref{appendix:datasets}.

\paragraph{Evaluation}
We evaluate the performance of models on the test sets of those three datasets and set the beam size to 4 in inference.
We employ SacreBLEU \cite{bleu, sacrebleu} to evaluate the quality of inferences at the word level and report BERTScore \cite{bertscore} of inferences at the representation level.
We measure the off-target ratio on zero-shot translations as a supplement.
We also conduct statistical significance testing \cite{significance}.
We describe our motivation in selecting evaluation metrics, the evaluation details, and the implementation of statistical significance testing in Appendix \ref{appendix:metric}.

\paragraph{Models}
When training from scratch, we implement a Transformer model with 6 encoder and decoder layers.
Given that those three datasets have different sizes, we set different hyper-parameters in training.
Then, three open-source models\footnote{Note that, those models are trained by adding a source language tag at the encoder and a target language tag at the decoder. In fine-tuning, we keep the original strategy.} are utilized in fine-tuning experiments, including M2M-418M, M2M-1.2B \cite{m2m} and mBART50 \cite{mbart50}.
The hyper-parameter settings can be found in Appendix \ref{appendix:models}.
Additionally, hyper-parameters are selected based on the ablation studies conducted on the validation sets, which is reported in Appendix \ref{appendix:ablation}.

\paragraph{Baselines}
Vanilla Transformer \cite{transformer, GooglesMNMT} is one of the baselines, denoted by \textsc{Vanilla} in the experiments of training from scratch.
Then, the baseline in fine-tuning experiments is the full-parameter fine-tuning, denoted by \textsc{F.T.}.
Moreover, three representative methods are reproduced in our experiments of training from scratch, the standard of baseline selection is as follows:
\begin{itemize}
\item  \textsc{SemAli}: \citet{Constras-2021} think the encoder output is language-agnostic, so they align the semantic information across different languages at the output of the encoder.
However, our analysis shows that this viewpoint is inaccurate because the semantic information is aligned by the subspace of the target language instead of the real language-agnostic.
When there are not any additional parameters introduced, \textsc{SemAli} still is the de-facto SOTA based on regularizing representations in MNMT.
\item  \textsc{DisPos}: \citet{DisentPos-2021} have the same objective as LoLE, however, they suggest reducing the constraint on the encoder \cite{gu-2019} by removing the residual connection, which is a different style that corresponds to the idea of biasing we used in LoLE.
\item \textsc{TLP}: \citet{TLP-2021} aim to add a loss to predict the language id at the top layer of the decoder, which is contrary to LCLR and our analysis in Section \ref{section:decoder} where we argue that the bottom layers of the decoder are more sensitive to the language features.
\end{itemize}

\subsection{Results}\label{section:results}
First of all, \citet{gu-2019,DisentPos-2021} pointed out that the vanilla Transformer is superior in supervised translation directions, i.e., en<->x, because the model excessively focuses on English, which is the language dominating the training set, to lose its generalization on non-English languages, i.e., the zero-shot translation.
Moreover, \citet{pareto_2023, pareto_nips} showed that improving the zero-shot may come at the expense of supervised performance.
In this work, our methods significantly improve the zero-shot translation without degrading the supervised performance in both training from scratch and fine-tuning.

Table \ref{tab:1} shows the experimental results of training from scratch.
In supervised translations of Europarl-15/TED-19/OPUS-100, \textsc{LoLE} shows divergent results of 0.13/-0.04/-0.22 on \texttt{en}$\to$\texttt{x} and 0.11/0.05/-0.02 on \texttt{x}$\to$\texttt{en}.
Similarly, \textsc{LCLR} shows diverse results of -0.05/-0.01/-0.03 and 0.04/-0.01/0.07, respectively.
Then, \textsc{Both} achieves the results of 0.18/-0.02/0.03 on \texttt{en}$\to$\texttt{x} and 0.12/0.12/-0.03 on \texttt{x}$\to$\texttt{en}.
In zero-shot translations, \textsc{BOTH} outperforms \textsc{Vanilla} 1.55/1.33/2.93 for BLEU and 0.58/1.02/3.12 for BERTScore.
Our models perform best in zero-shot translations of Europarl-15 and OPUS-100, and the improvements in zero-shot translations are always statistically significant.
Note that, although \textsc{SemAli} achieves the best zero-shot translation performance in TED-19, the supervised performance of \textsc{SemAli} is significantly degraded compared to \textsc{Vanilla}, which is a common and unresolved problem \cite{gu-2019,massive-2020, DisentPos-2021}.
On the contrary, our methods not only perform competitively with \textsc{SemAli} in zero-shot translations but also benefit the supervised translation capacity.
Moreover, these two proposed methods are orthogonal, which can be proved by assessing \textsc{LoLE}, \textsc{LCLR} and \textsc{Both} individually: (1) \textsc{LoLE} achieves gains of 1.53/1.22/2.85 for BLEU and 0.54/0.91/3.34 for BERTScore; (2) \textsc{LCLR} improves 1.06/0.14/0.09 and 0.39/0.13/0.06 scores; (3) The gains of \textsc{Both} are always higher than \textsc{LoLE} and \textsc{LCLR}.
In addition, we can observe that the improvement of \textsc{LCLR} is limited in TED-19 and OPUS-100, which can be attributed to the diverse languages involving in these two datasets and being easily distinguished by the vanilla decoder.
This result also supports that the main challenge of MNMT is the transfer within the encoder.
Thus, we can conclude that our methods substantially benefit the zero-shot translation capacity of MNMT models.

\begin{table}[t]
  \centering
  \resizebox{0.5\textwidth}{!}{
    \begin{tabular}{l cccccc}
    \toprule
    & \multicolumn{3}{c}{BLEU} & \multicolumn{3}{c}{BERTScore} \\
    \midrule
    Method & \texttt{en}$\to$ & $\to$\texttt{en} & zero.* & \texttt{en}$\to$ & $\to$\texttt{en} & zero* \\
    \midrule
    M2M-418M & 21.88 & 26.43 & 14.51 & 82.52 & 93.25 & 79.26 \\
    \hspace{3mm}\textsc{F.T.} & 26.68 & 32.95 & 17.46 & 84.47 & 94.30 & 80.79 \\
    \hspace{3mm}\textsc{LoLE} & 26.81  & 33.16  & 17.52  & \textbf{84.51} & 94.31  & 80.84 \\
    \hspace{3mm}\textsc{LCLR} & 26.81  & \hspace{0.4em}\textbf{33.67}$^\dag$  & 17.65  & 84.47  & \textbf{94.40}  & 80.88 \\
    \hspace{3mm}\textsc{BOTH} & \textbf{26.83} & \hspace{0.4em}33.63$^\dag$  & \textbf{17.68}  & 84.49  & 94.38  & \textbf{80.90} \\
    \midrule
    M2M-1.2B & 24.32& 28.94& 15.95& 83.17& 93.72& 79.75 \\
    \hspace{3mm}\textsc{F.T.} & 27.71& \textbf{34.97}& 18.48& 84.71& \textbf{94.53} & 81.14 \\
    \hspace{3mm}\textsc{LoLE} & \hspace{0.4em}28.29$^\dag$ & 34.12 & 18.67 & \hspace{0.4em}84.91$^\dag$ & 94.48 & \hspace{0.4em}\textbf{81.26}$^\dag$ \\
    \hspace{3mm}\textsc{LCLR} & \hspace{0.4em}28.26$^\dag$ & 34.54 & 18.64 & \hspace{0.4em}84.90$^\dag$ & 94.50 & 81.22 \\
    \hspace{3mm}\textsc{BOTH} & \hspace{0.4em}\textbf{28.37}$^\dag$ & 34.59 & \textbf{18.69} & \hspace{0.4em}\textbf{84.92}$^\dag$ & 94.51 & 81.23 \\
    \midrule
    mBART50 & 25.28 & 33.50 & 6.92 & 83.93 & \textbf{94.43} & 72.91 \\
    \hspace{3mm}\textsc{F.T.} & 27.17 & 33.96 & 5.58 & 84.64 & 94.36 & 72.96 \\
    \hspace{3mm}\textsc{LoLE} & 27.19 & 33.93 & \hspace{0.4em}7.28$^\dag$ & 84.60 & 94.37 & \hspace{0.4em}73.86$^\dag$ \\
    \hspace{3mm}\textsc{LCLR} & 27.07 & 34.02 & \hspace{0.4em}\textbf{9.69}$^\dag$ & 84.59 & 94.38 & \hspace{0.4em}75.31$^\dag$ \\
    \hspace{3mm}\textsc{BOTH} & \textbf{27.36}& \textbf{34.04}& \hspace{0.4em}9.55$^\dag$& \textbf{84.66} & 94.36 & \hspace{0.4em}\textbf{75.55}$^\dag$ \\
    \bottomrule
    \end{tabular}
  }
  \caption{Averaged scores for experiments of fine-tuning.
           \textsc{F.T.} means fine-tuning without any trick.
           * is added to zero. to show it is not a real zero-shot scenario for M2M.
           The bold number indicates the best result, and the numbers with $\dag$ are significantly better than \textsc{F.T.}.
           The off-target ratios are reported in Appendix \ref{appendix:offratio}.
  }
  \label{tab:2}
\end{table}
Table \ref{tab:2} shows the experimental results of fine-tuning.
For M2M-418M, compared with \textsc{F.T.}, our methods obtain up to 0.15/0.72/0.22 for BLEU scores and 0.04/0.10/0.11 for BERTScore in \texttt{en}$\to$\texttt{x}, \texttt{x}$\to$\texttt{en} and zero-shot translations, respectively;
For M2M-1.2B, the gain is up to 0.66/-0.38/0.21 for BLEU scores and 0.21/-0.02/0.12 for BERTScore;
For mBART50, the gain is up to 0.19/0.08/4.11 for BLEU scores and 0.02/0.02/1.69 for BERTScore.
Those scores show that the improvement on M2M is marginal compared with training from scratch.
This derives M2M is trained by interconnected translation pairs instead of an English-centric dataset, which results in the robust transferability of multilingual representations.
However, the degeneration on \textsc{F.T.} of mBART50 shows that fine-tuning drastically influences the zero-shot translation capacity.
For instance, the BLUE scores of \texttt{fr}$\to$\texttt{vi} decrease to 11.84 from 20.57 and \texttt{fr}$\to$\texttt{zh} increase to 13.52 from 1.90, but our model obtains 18.47 and 17.19, respectively.
Such results and the significant testing indicate again the advantage of our proposed methods in improving multilingual representations for zero-shot translation capacity.

\section{Discussion}
\begin{figure}[!t]
    \centering
    \begin{subfigure}[b]{0.48\linewidth}
        \centering
        \includegraphics[width=\linewidth]{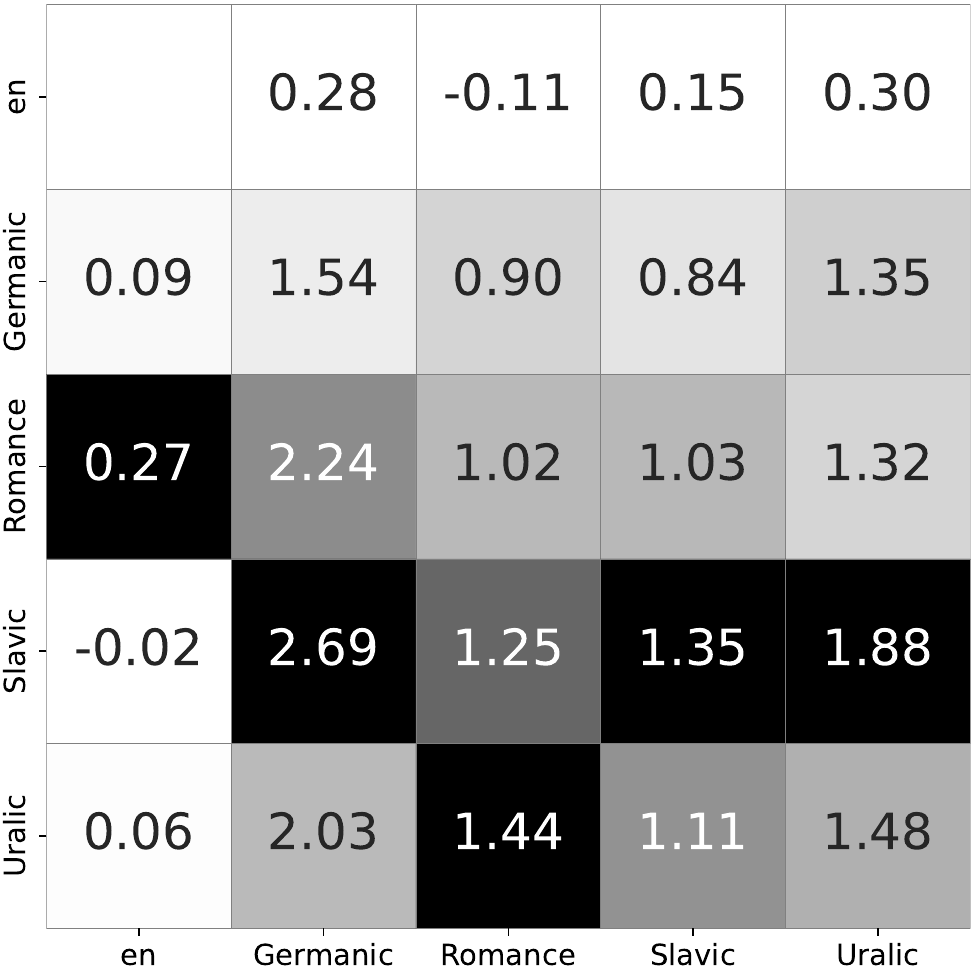}
        \caption{BLEU}
        \label{fig:correlation_1}
    \end{subfigure}
    \begin{subfigure}[b]{0.48\linewidth}
        \centering
        \includegraphics[width=\linewidth]{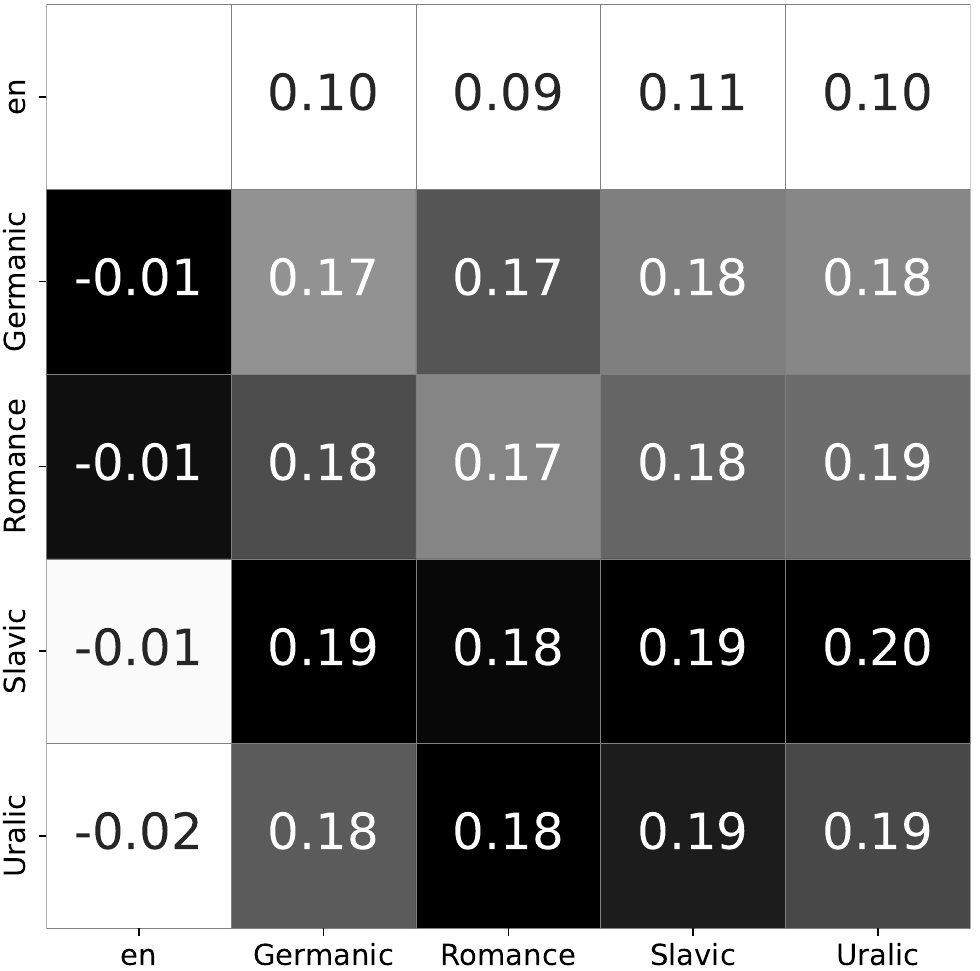}
        \caption{SVCCA}
        \label{fig:correlation_2}
    \end{subfigure}
    \caption{Differences between our model and \textsc{Vanilla}. X-axis is the target language family where \texttt{en} is considered solely. Hence, we plot the color ladder by column where the darker the color, the bigger the difference.}
    \label{fig:correlation}
\end{figure}
\subsection{Correlation between Representational Disentanglements and Improvements}\label{section:correlation}
Table \ref{tab:1} shows the overall results by taking averages across all language pairs, which may overlook pair-specific tendencies.
Therefore, we group Europarl-15 by the language families and report the average scores of translating from one language family to another.
Figure \ref{fig:correlation_1} shows the difference in BLEU scores between our models and \textsc{Vanilla}.
As shown in Figure \ref{fig:correlation_2}, we also compute the SVCCA scores between the identity of the non-central language and the identity of the central language at the encoder's output and group them in the same manner.
Given the similar distribution in Figure \ref{fig:correlation}, we conduct Pearson correlation analysis \cite{Pearson} of all language pairs instead of language families in Europarl-15, and we compute the coefficients and $p$-values of Pearson correlation by target languages to maintain fairness.
We observe two key points:
1) The coefficient and $p$-value of \texttt{en} are -0.087 and 0.76, respectively.
This result suggests that there is no statistical correlation, which is predictable because \texttt{x}$\to$\texttt{en} is not affected by representational entanglements.
2) The coefficient and $p$-value of non-central languages are in the ranges of 0.585 to 0.855 and 4e-5 to 0.021, respectively.
In more detail, the mean values are 0.770 and 0.002 and the variances are 0.04 and 3e-5, respectively.
This analysis proves that the degree of representational disentanglement positively correlates with the improvement of zero-shot translations.

\subsection{Analysis of Improved Representation}\label{section:improve}
\begin{figure}[!t]
    \centering
    \begin{subfigure}[b]{0.48\linewidth}
        \centering
        \includegraphics[width=\linewidth]{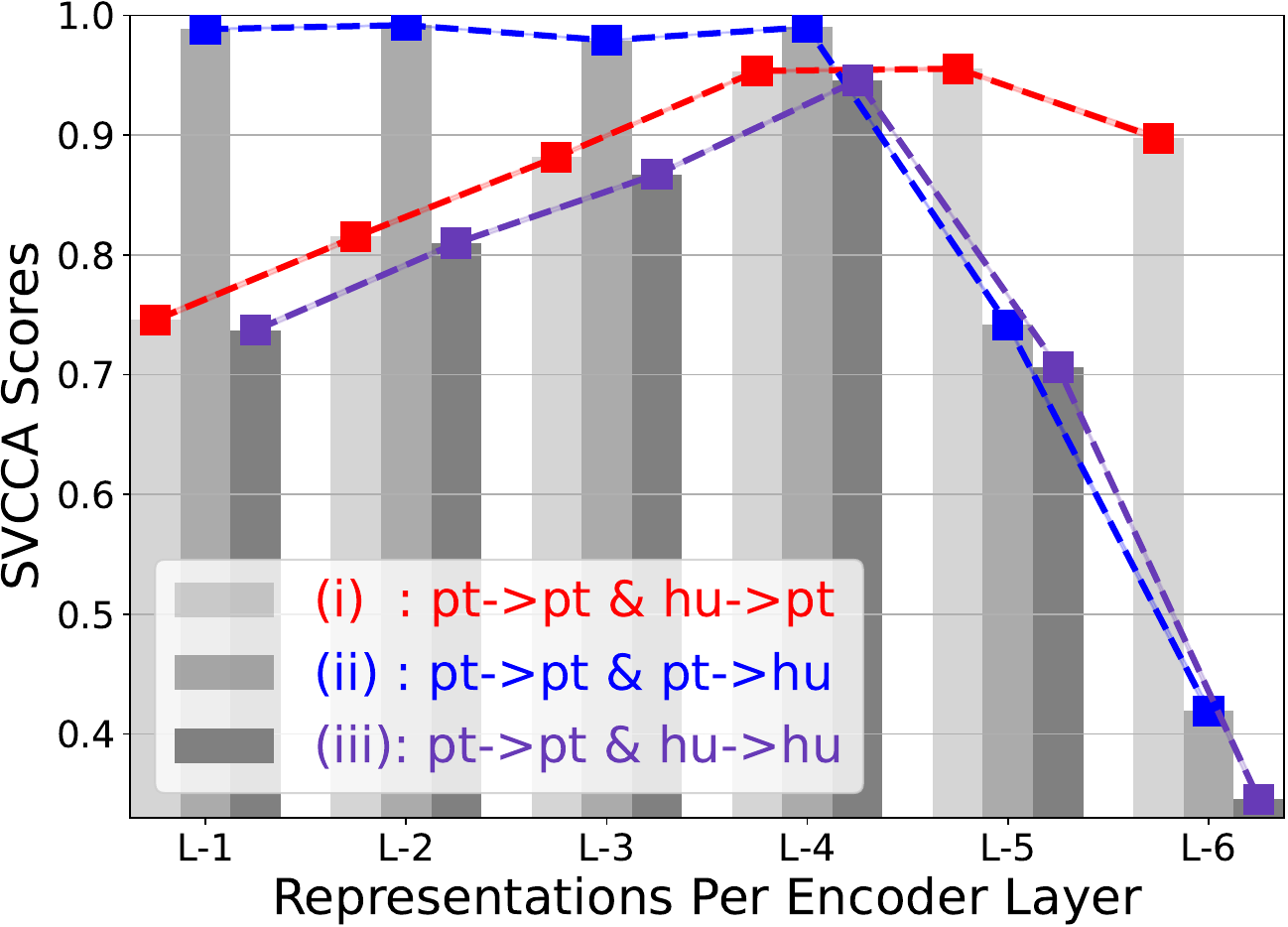}
        \caption{Compare to Fig. \ref{fig:transfer_2}.}
        \label{fig:analysis_1}
    \end{subfigure}
    \begin{subfigure}[b]{0.48\linewidth}
        \centering
        \includegraphics[width=\linewidth]{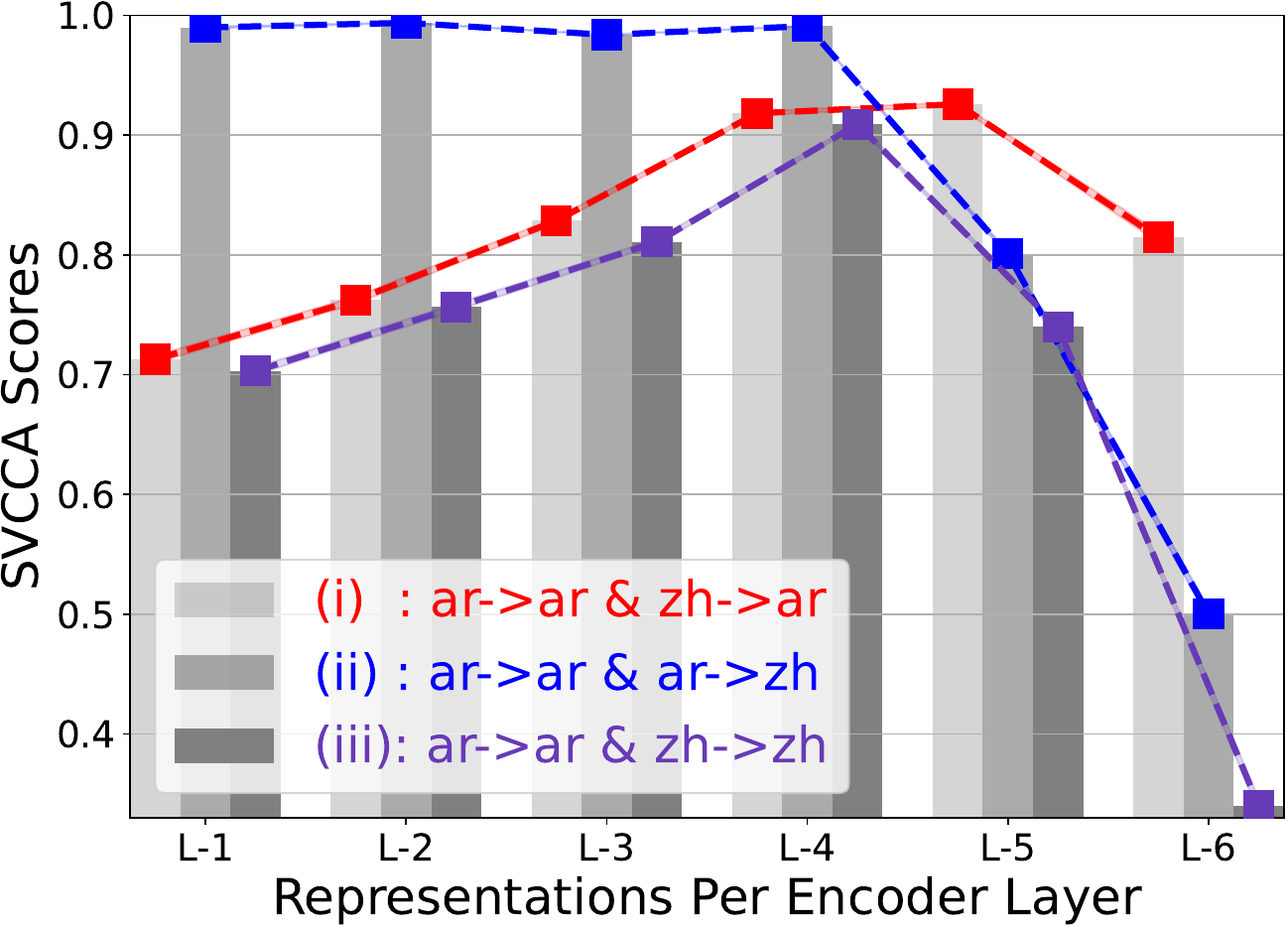}
        \caption{Compare to Fig. \ref{fig:transfer_4}.}
        \label{fig:analysis_2}
    \end{subfigure}
    \begin{subfigure}[b]{0.48\linewidth}
        \centering
        \includegraphics[width=\linewidth]{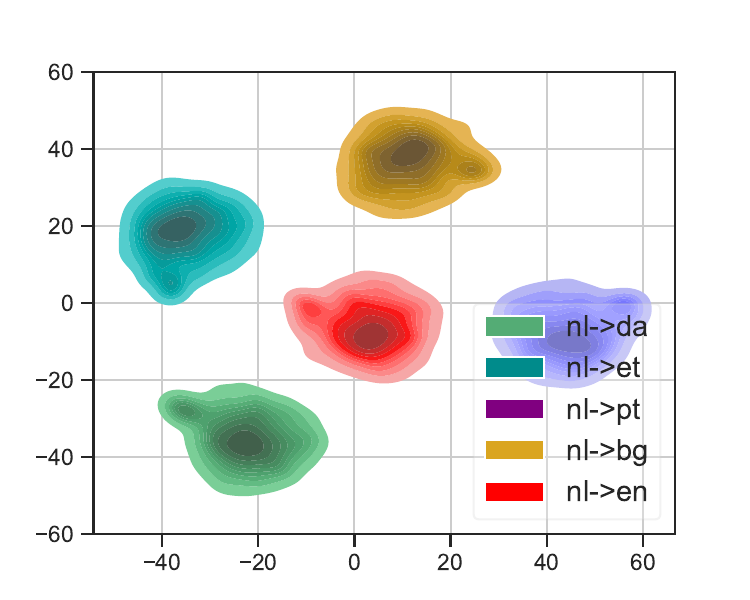}
        \caption{Compare to Fig. \ref{fig:entangle_5}.}
        \label{fig:analysis_3}
    \end{subfigure}
    \begin{subfigure}[b]{0.48\linewidth}
        \centering
        \includegraphics[width=\linewidth]{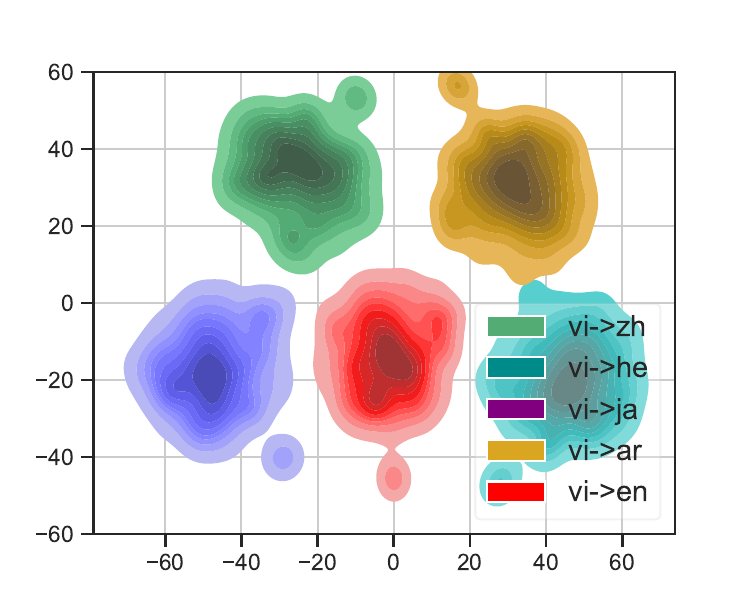}
        \caption{Compare to Fig. \ref{fig:entangle_6}.}
        \label{fig:analysis_4}
    \end{subfigure}
    \begin{subfigure}[b]{0.48\linewidth}
        \centering
        \includegraphics[width=\linewidth]{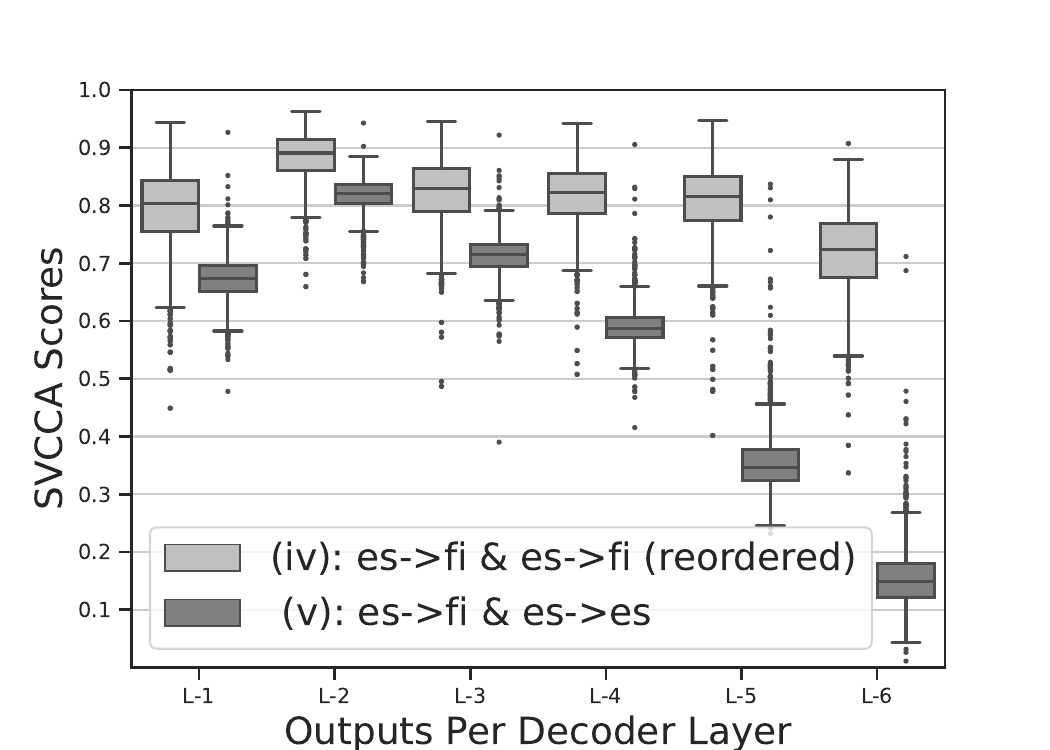}
        \caption{Compare to Fig. \ref{fig:decoder_1}.}
        \label{fig:analysis_5}
    \end{subfigure}
    \begin{subfigure}[b]{0.48\linewidth}
        \centering
        \includegraphics[width=\linewidth]{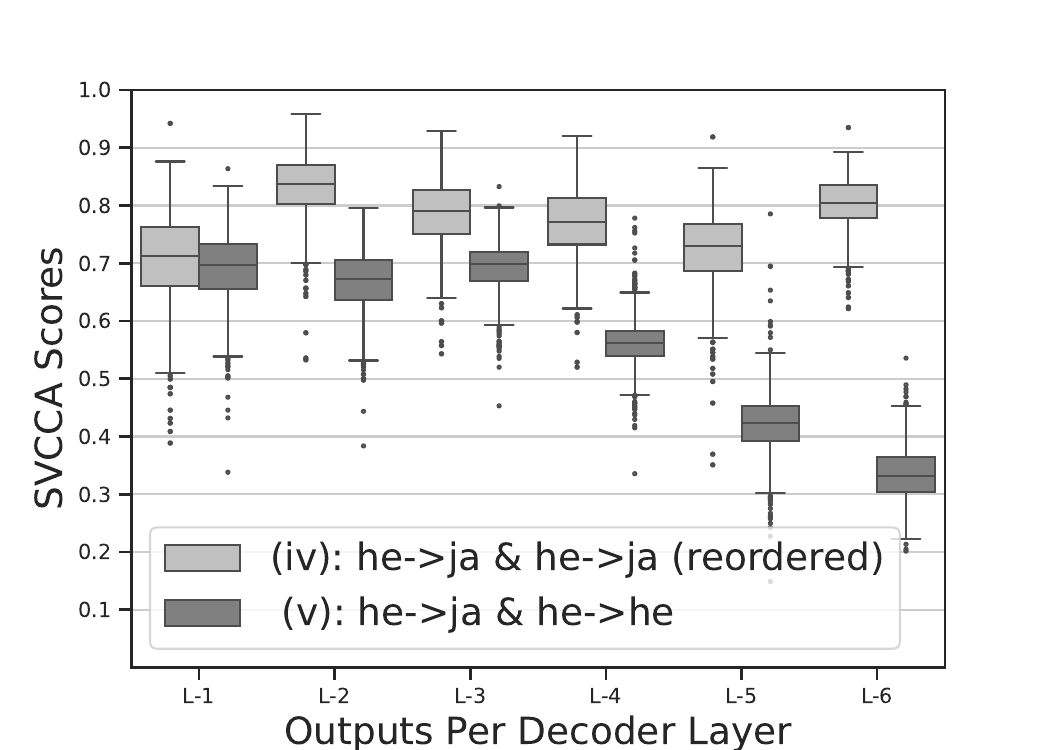}
        \caption{Compare to Fig. \ref{fig:decoder_2}}
        \label{fig:analysis_6}
    \end{subfigure}
    \caption{Visualizations for the encoder that incorporates \textsc{LoLE} and \textsc{LCLR}, showing improvements compared with \textsc{Vanilla}.
    Additionally, the model plotted in \ref{fig:analysis_5} only incorporates \textsc{LCLR}.}
    \label{fig:analysis}
\end{figure}
We measure representation transfer in the model incorporating our proposed methods to verify our findings further.
As shown in Figures \ref{fig:analysis_1} and \ref{fig:analysis_2}, both scenarios exhibit improvements on (\romannumeral1).
Meanwhile, Figures \ref{fig:analysis_3} and \ref{fig:analysis_4} indicate that the entanglement of representations among languages at the encoder is resolved.
The evidence suggests that \textsc{LoLE} effectively enhances representation transfer in the encoder.
Additionally, (\romannumeral2) and (\romannumeral3) in Figures \ref{fig:analysis_1} and \ref{fig:analysis_2} also achieve higher scores at lower layers of the encoder, which suggests that \textsc{LoLE} indeed makes lower layers of the encoder focus on surface-level information.
By contrast, as shown in Figures \ref{fig:analysis_5} and \ref{fig:analysis_6}, the more stable trend of (\romannumeral4) in both scenarios suggests that \textsc{LCLR} can improve the capacity of lower layers of the decoder to distinguish languages to improve zero-shot translations.
In addition, Appendix \ref{appendix:improve} provides the representational analysis for fine-tuning models, which proves that target language features are consistently beneficial in the encoder.

\section{Related Works}
Prior studies on analyzing multilingual representation in Section \ref{section:priorworks} led to several effective methods in MNMT.
Some works focused on updating and constraining the encoder to improve multilingual representations, and the findings in discrepancy mentioned in Section \ref{section:priorworks} led to two distinct approaches.
First, \citet{Constras-2021, Regular-2022, Regular-2023, aganostic-2024} suggested regularizing the encoder for aligning semantic information across different source languages by introducing additional training objectives.
Similarly, \citet{constrainEncoder, interlingua-2020} explicitly modified the output form of the encoder to transfer the representation of the source sentence toward a language-agnostic state.
Second, \citet{gu-2019, DisentPos-2021, lcs-2024} introduced specialized modeling constraints to improve the encoder to transfer source sentence representations to the target language without adding extra parameters, and \citet{share-2021, lsls-2023} enhanced the representation of target language information by simply adding language-specific modules.
Additionally, \citet{TLP-2021, adaptingZero, aganostic-2024} focused on improving the target language representation on the decoder side or adding modules specified to the target language to the decoder.
Given that the above works can all be encompassed within our analyses, we argue that this work offers insights for future improvements in MNMT. Specifically, enhancing the encoder to transfer source language representations into the target language subspace and align semantic information within those subspaces is the key to improving MNMT.

In addition, a critical factor of this work is the introduction of the identity pair as an analytical tool.
Specifically, while identity pairs have been heuristically used in prior works \cite{measuring-2019,automatic-2020,aganostic-2024}, as an assumed indicator of language-specific representation states, they have not been subject to systematic or quantitative analysis.
In contrast, we explicitly define, validate, and utilize identity pairs to probe representational properties in a controlled and measurable way. This not only strengthens the empirical basis of our conclusions but also constitutes an important methodological contribution of this work.

\section{Conclusion}
We systematically investigated the representational issue of zero-shot translation deficiency in multilingual neural machine translation models.
Our analyses show that the encoder transfers translation representations from the source language to the target language, and aligns semantics across different source languages at the target language subspace.
We applied engineering practices to verify our findings by proposing two orthogonal methods, which substantially improve the zero-shot translation capacity.
Thus, our findings are significant for guiding the improvement of the transferability of multilingual representations.

\section{Limitations}

\begin{figure}[h]
    \centering
    \begin{subfigure}[b]{0.48\linewidth}
        \centering
        \includegraphics[width=\linewidth]{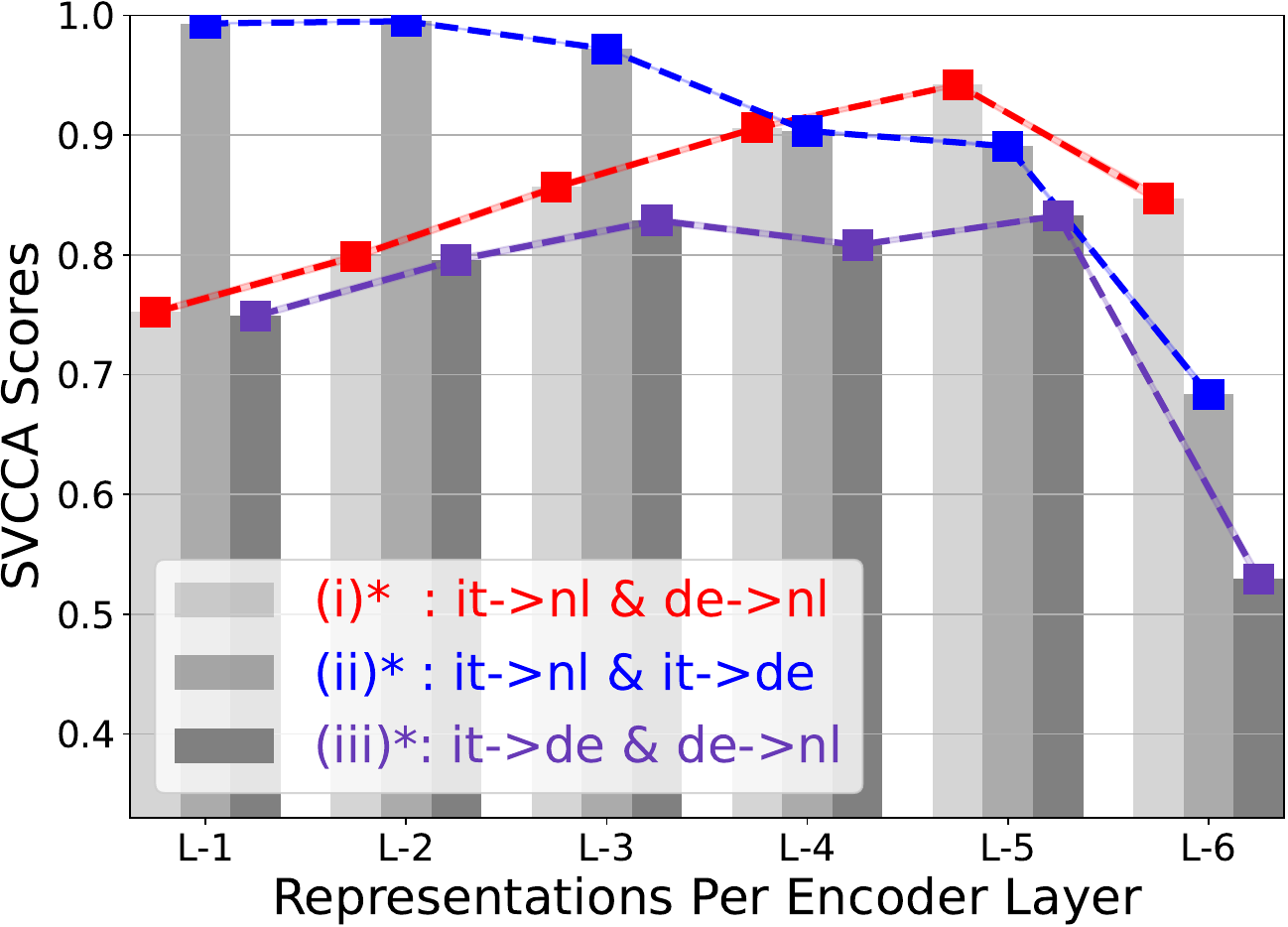}
        \caption{Bidirection}
        \label{fig:limitation_1}
    \end{subfigure}
    \begin{subfigure}[b]{0.48\linewidth}
        \centering
        \includegraphics[width=\linewidth]{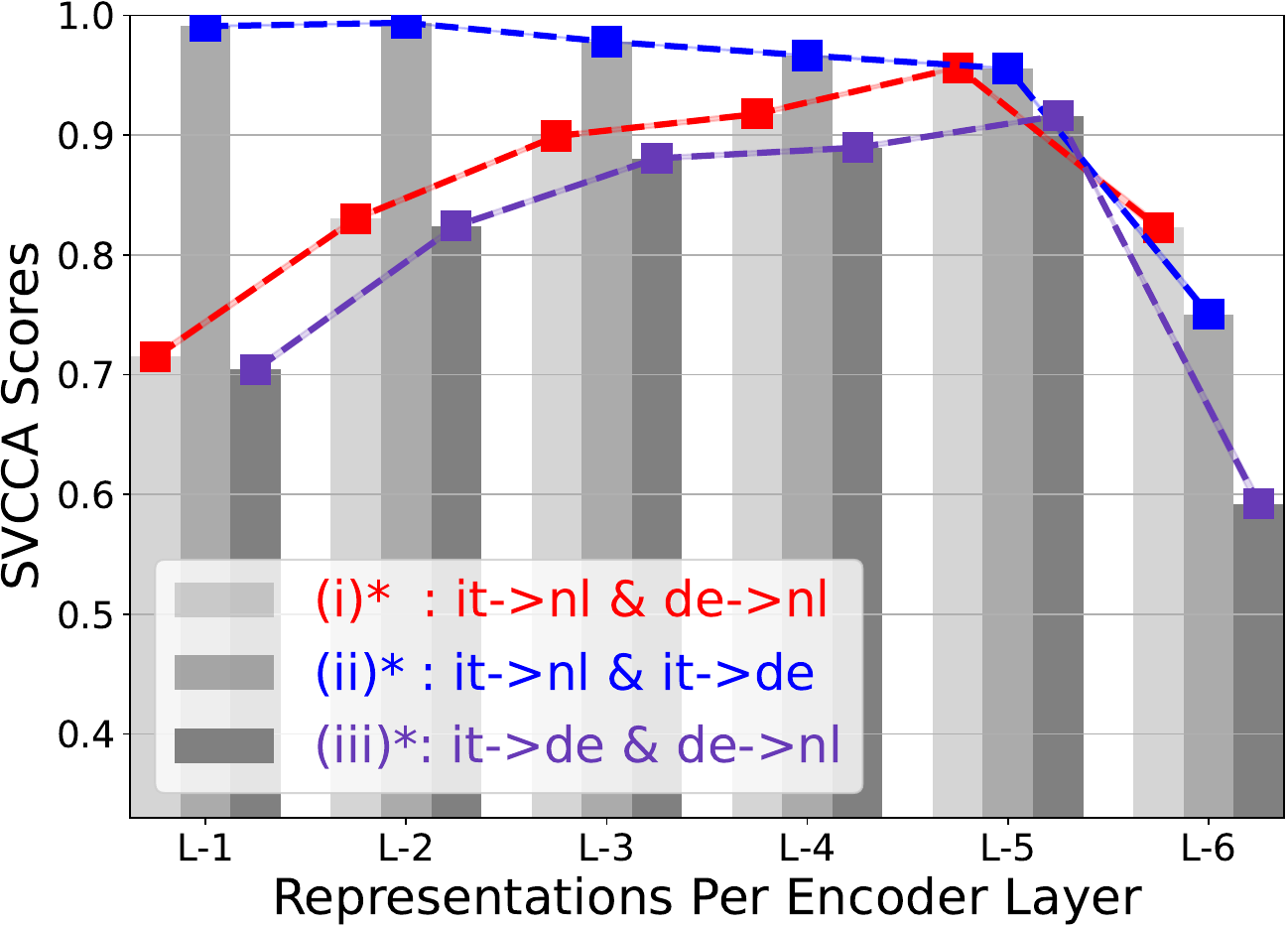}
        \caption{Unidirection}
        \label{fig:limitation_2}
    \end{subfigure}
    \caption{Illustration of the comparison between the bidirectional and the unidirectional scenarios.
    \ref{fig:limitation_1} has the same model settings with Figure \ref{fig:transfer}, but analyzes the same pairs with \ref{fig:limitation_2}.
    }
    \label{fig:limitation}
\end{figure}
This work has two limitations.
First, the identity pair is a proxy of language representations based on bi-directional training, i.e., each non-central language appears in the encoder and decoder together.
Therefore, we designed an additional study to investigate the impact by retraining a model by eliminating \texttt{nl}$\to$\texttt{en} and \texttt{en}$\to$\texttt{it} so that \texttt{it} and \texttt{nl} appear only in the encoder and decoder, respectively, based on the analysis in Section \ref{section:transfer}.
Then, we conducted a comparison by taking \texttt{de} as the middle language to perform the role of identity pairs in analysis.
As shown in Figure \ref{fig:limitation}, the target language features keep the same trend as shown in Section \ref{section:transfer} to support our conclusion again, but the influence of source language features increases relatively.

The second limitation is our investigation is based on adding a language tag specified to the target language at the beginning of the source sentence for the encoder.
Although this is the de facto MNMT training strategy \cite{GooglesMNMT,aharoni-2019, arivazhagan-2019, gu-2019,constrainEncoder,TagMatter_2021,TLP-2021,Constras-2021,adaptingZero, target-off,Regular-2022, Regular-2023}, the current open-source models \cite{m2m, mbart50, nllb} are based on another strategy, i.e., adding a source language tag at the encoder side and adding a target language tag at the decoder side.
Although, in Section \ref{section:experiments}, we have shown our proposed methods also benefit models with this strategy, this effectiveness is proved by empirical experiments.
Thus, our future work is to investigate the representation transfer of this strategy to guide further improvements.

\section{Further Considerations}
\paragraph{Ethical Consideration}
All datasets used in this work are public data, which are proven harmless.
Moreover, this work is foundational research and is not tied to particular applications.
Thus, there is no ethical risk existed in this work.
\paragraph{Sustainability statement}
As noted in Appendix \ref{appendix:models}, the GPU used in training individual models is A6000, which has an estimated carbon dioxide emission of approximately 0.13 kg per hour.\footnote{Measured by \url{https://mlco2.github.io/impact/}.}
Specifically, models trained on Europarl-15 and TED-19 required approximately 48 GPU hours, while models trained on OPUS-100 necessitated around 192 GPU hours.

\bibliography{mtsummit25}

\appendix

\section{Sentence-level SVCCA Score}\label{appendix:svcca}
We use SVCCA \cite{svcca} to measure representation similarity in MNMT \cite{investigateMNMT}.
We follow the approach of \citet{DisentPos-2021} so that similarity is measured at the sentence level to ensure that each score is computed on equivalent features without the influence of other sentences in the set.

Based on the definition of Section \ref{section:mnmt}, we denote hidden representations of a sentence by $\mathbf{H} = \{\boldsymbol{h}_1, \boldsymbol{h}_2 ... \boldsymbol{h}_q$\}, where  $\mathbf{H} \in \mathbb{R}^{q \times d}$, $q$ equals to the length $n$ or $m$ from either the encoder or decoder, and $d$ is the model dimension.
Additionally, the practical length is $n+1$ when $\mathbf{H}$ is fed into the encoder because the encoder receives the input concatenated by $l$ and $\boldsymbol{x}$\footnote{$l$ plays the role of translation instruction instead of a token belonging to the target language with semantics, thus, this concatenation would not influence the measurement by mixing target language information into the sentence representation within the encoder.}.
Then, we derive the sentence-level representation $\overline{\boldsymbol{h}}$ using average pooling 
$\overline{\boldsymbol{h}} = \frac{\sum_{i=1}^q \boldsymbol{h}_i}{q}$
, which mainly represents the language features and semantics of the source sentence rather than syntactic information because positional information is reduced.

Given $\mathbf{H}^a$ and $\mathbf{H}^b$ derived from two sentences, SVCCA first performs singular value decomposition on their averaged representations to obtain subspace representations  ${\overline{\boldsymbol{h}}^{a}} \in \mathbb{R}^{d^a}$ and $ {\overline{\boldsymbol{h}}^{b}}  \in \mathbb{R}^{d^b}$, where noise is reduced \cite{svccallm-2019}.
Then we perform canonical correlation analysis \cite{cca} to determine $\mathbf{W}^a \in \mathbb{R}^{d' \times d^a}$ and $\mathbf{W}^b \in  \mathbb{R}^{d' \times d^b}$.
Formally, we compute correlation $\rho$ between $\overline{\boldsymbol{h}}^a$ and $\overline{\boldsymbol{h}}^b$ as
\begin{equation}
\label{eq2}
\rho =
\frac{
\langle 
\mathbf{W}^a {\overline{\boldsymbol{h}}^a},
\mathbf{W}^b {\overline{\boldsymbol{h}}^b} \rangle}
{\Vert 
\mathbf{W}^a {\overline{\boldsymbol{h}}^a} 
\Vert \Vert
\mathbf{W}^b {\overline{\boldsymbol{h}}^b} 
\Vert},
\end{equation}
where $\langle \cdot, \cdot \rangle$ indicates the inner product. We use $\rho$ to represent the similarity of two sentences.
Finally, we compute the set-level score by taking the average scores of all sentences over the set.

\section{Detailed Information of Datasets}\label{appendix:datasets}
This work involves three datasets, i.e., Europarl-15, TED-19, and OPUS-100, where Europarl-15 and TED-19 are used in preliminary experiments.
The training sets of those three datasets have different sizes, but the validation and test sets of a pair generally contain 2,000 translation instances.
In preliminary experiments, we measure SVCCA scores in the test sets because those instances are unseen in the training.

Europarl-15 is collected from MMCR4NLP, which has high-quality translation instances and each instance in a language is one-to-one corresponding to other languages, i.e., all language-specific sets have parallel semantics \cite{europarl, mmcr4nlp}, including 15 European languages from 4 language families.
Specifically, Germanic includes \texttt{en}, \texttt{de}, \texttt{nl}, \texttt{da}, Romance includes \texttt{es}, \texttt{pt}, \texttt{it}, \texttt{ro}, Slavic includes \texttt{sl}, \texttt{bg}, \texttt{pl}, \texttt{cs}, and Uralic includes \texttt{fi}, \texttt{et}, \texttt{hu}.
The training and validation sets cover 28 supervised translation pairs where English is the central language used to bridge the non-central languages.
The test set consists of all language pairs, including 182 zero-shot translation pairs in addition to supervised translation pairs.
Finally, each pair in the training set comprises 189,310 instances.

In contrast to Europarl-15, which is the semantically aligned dataset, TED-19 consists of 19 languages, including \texttt{en}, \texttt{ar}, \texttt{he}, \texttt{ru}, \texttt{ko}, \texttt{it}, \texttt{ja}, \texttt{zh}, \texttt{es}, \texttt{nl}, \texttt{vi}, \texttt{tr}, \texttt{fr}, \texttt{pl}, \texttt{ro}, \texttt{fa}, \texttt{hr}, \texttt{cs}, \texttt{de}, which belong to various language families without parallel semantics, from TED Talks \cite{ted}.
Each translation pair contains 103,093 to 214,111 instances in training, and the training set comprises 6,551,456 instances in total.
Because of the unparallel semantics of TED-19, we align \texttt{ar}, \texttt{he}, \texttt{zh}, \texttt{hr}, \texttt{vi}, \texttt{ja} to obtain 967 translation instances for measuring SVCCA scores.
In addition, the reason why the number of languages is 19 is that, first, TED Talks have 20 high-resource languages, which are supported in M2M \cite{m2m} and mBART50 \cite{mbart50}.
However, the tokenization of \texttt{th} is problematic, resulting in deprecating \texttt{th}.

OPUS-100 consists of 95 languages, 188 pairs, and 109.2 million instances in total \cite{massive-2020,TLP-2021}, where 90 pairs comprise 1 million instances and 56 pairs have more than 0.1 million instances.
Different from \citet{TLP-2021}, we do not include the zero-shot translation pairs in the validation set to avoid biases when assessing the transferability of multilingual representations.

\section{Detailed Settings of Models}\label{appendix:models}
We implement the Transformer \cite{transformer} as the backbone model via Fairseq \cite{fairseq}.
For the configuration of models trained on Europarl-15 and TED-19, we follow \citet{investigateMNMT} to set 6 encoder and decoder layers.
Based on the ablation study conducted in the validation set in Europarl-15 and TED-19 shown in Appendix \ref{appendix:ablation}, we apply \textsc{LoLE} in the fifth encoder layer, set $d^e$ to 128, and set $d^h$ and $k$ of \textsc{LCLR} to 64 and 30, respectively.
When we solely apply \textsc{LCLR}, we set the position to the bottom decoder layer based on the findings in Section \ref{section:decoder}.
When we integrate both \textsc{LoLE} and \textsc{LCLR} into a model, we relocate \textsc{LCLR} to the second-bottom decoder layer because of the improved language features of the encoder representations.
We adopt a shared vocabulary trained by SentencePiece \cite{sentencepiece} with 50,000 tokens for both the encoder and decoder.
The model consists of 4 attention heads, embedding size of 512, inner size of 1024, dropout rate of 0.2, maximum learning rate of 0.0005 with the inverse square root schedule and 4,000 warmup steps, and label smoothing rate of 0.1.
We set the batch size to 8,000 tokens per GPU, apply Adam \cite{adam} as the optimizer, and set temperature sampling with $T=5$ \cite{temperatureSample}.
We train the model with 60 epochs for Europarl-15 and 30 epochs for TED-15, and finally average the top 5 checkpoints using the loss on the validation set.
Compared with the basic configuration, the models trained on OPUS-100 have 8 attention heads, embedding size of 512, inner size of 2048, dropout rate of 0.1, and shared vocabulary size of 64,000.
We enlarge $d^e$ to 256 and $d^h$ to 128 for models trained on OPUS-100 and three pre-trained models because they involve more languages.
We train the model of OPUS-100 for 400,000 update steps with a batch size of 8,000 tokens per GPU for OPUS-100 and directly use the best checkpoint selected using the loss on the validation set.
Furthermore, models with Europarl-15 and TED-19 are trained on 8 NVIDIA V100 GPUs, and models with OPUS-100 are trained on 4 NVIDIA A6000 GPUs by setting \textit{--update-freq} to 2 in Fairseq to simulate 8 GPUs.

Three open-source models are utilized in fine-tuning experiments.
The first is M2M-418M \cite{m2m}, trained on standard multilingual translation tasks and supporting translation across 100 languages.
It is based on Transformer architecture, configured with 12 encoder and decoder layers, embedding size of 1024, inner size of 4096, and vocabulary size of 128,112, which results in a total of 418 million parameters.
The second model, M2M-1.2B \cite{m2m}, enlarges the number of layers to 24 and the inner size to 8192 on M2M-418M, and culminates in 1.2 billion parameters.
The last model is mBART50 \cite{mbart50}, trained on monolingual corpora across 50 languages following \citet{bart, mbart} and preliminarily fine-tuned for MNMT.
It shares the same parameter setup as M2M-418M with a vocabulary size of 250,053, which consists of 611 million parameters.
We conduct experiments on TED-19 because all covered languages are supported by these models.

\section{Detailed Introductions of Figure \ref{fig:comparison}}\label{appendix:introduction}
\begin{figure}[t]
    \centering
    \begin{subfigure}[b]{0.3\linewidth}
        \centering
        \includegraphics[width=\linewidth]{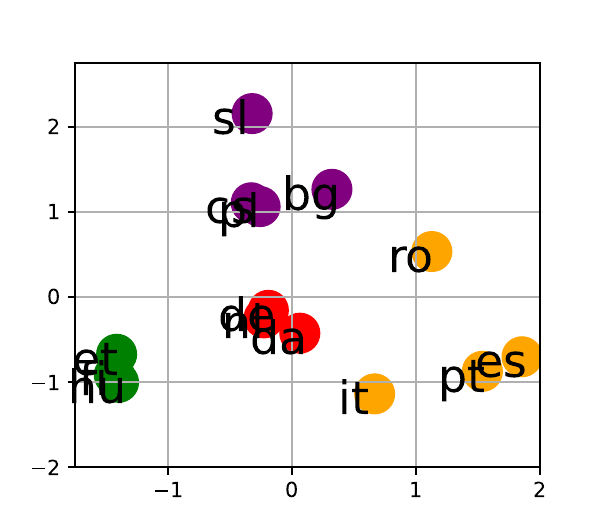}
        \caption{First layer}
        \label{fig:append_cluster_1}
    \end{subfigure}
    \begin{subfigure}[b]{0.3\linewidth}
        \centering
        \includegraphics[width=\linewidth]{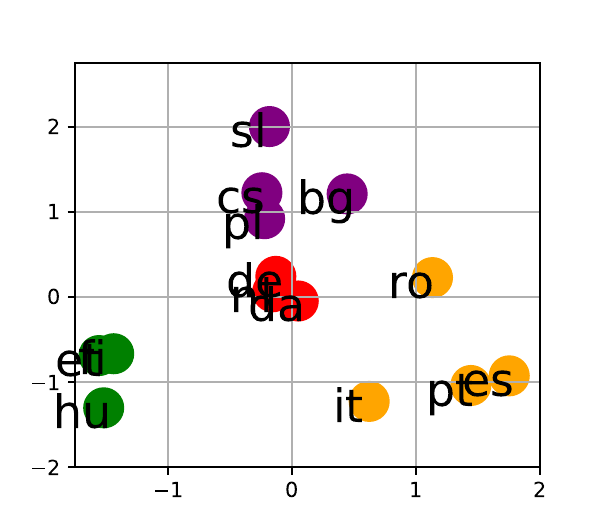}
        \caption{Second layer}
        \label{fig:append_cluster_2}
    \end{subfigure}
    \begin{subfigure}[b]{0.3\linewidth}
        \centering
        \includegraphics[width=\linewidth]{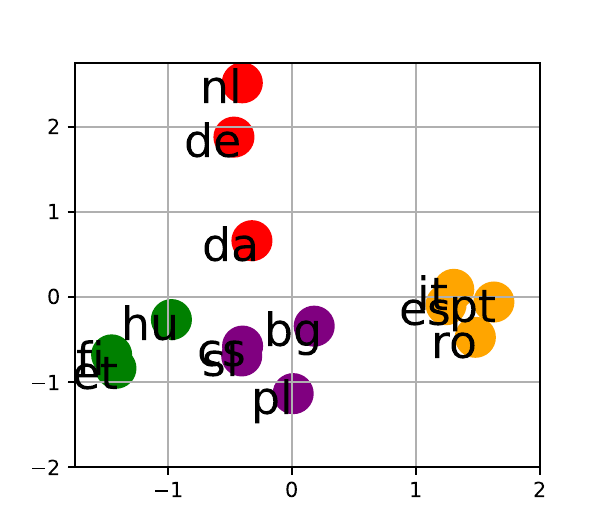}
        \caption{Third layer}
        \label{fig:append_cluster_3}
    \end{subfigure}
    \begin{subfigure}[b]{0.3\linewidth}
        \centering
        \includegraphics[width=\linewidth]{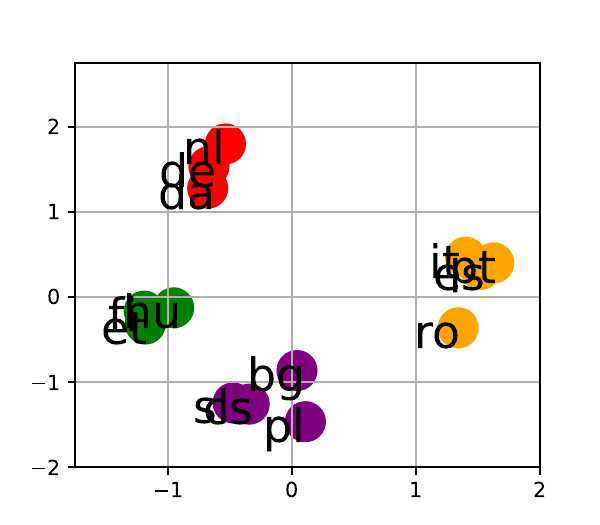}
        \caption{Fourth layer}
        \label{fig:append_cluster_4}
    \end{subfigure}
    \begin{subfigure}[b]{0.3\linewidth}
        \centering
        \includegraphics[width=\linewidth]{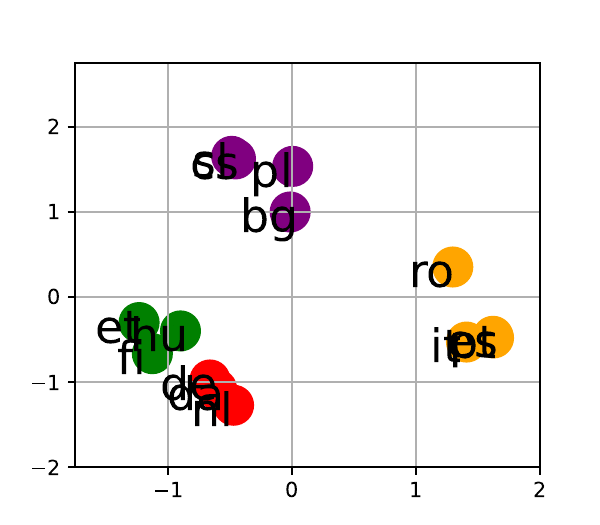}
        \caption{Fifth layer}
        \label{fig:append_cluster_5}
    \end{subfigure}
    \begin{subfigure}[b]{0.3\linewidth}
        \centering
        \includegraphics[width=\linewidth]{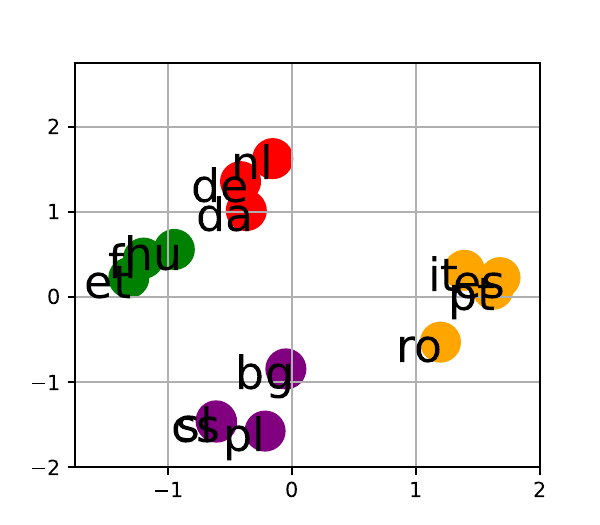}
        \caption{Sixth layer}
        \label{fig:append_cluster_6}
    \end{subfigure}
    \caption{Affinities for \texttt{en}$\to$\texttt{x} at each encoder layer. Language families of Europarl-15 are distinguished by colors: Germanic by red, Romance by yellow, Slavic by purple, and Uralic by green.}
    \label{fig:append_cluster}
\end{figure}

\begin{figure}[t]
    \centering
    \begin{subfigure}[b]{0.45\linewidth}
        \centering
        \includegraphics[width=\linewidth]{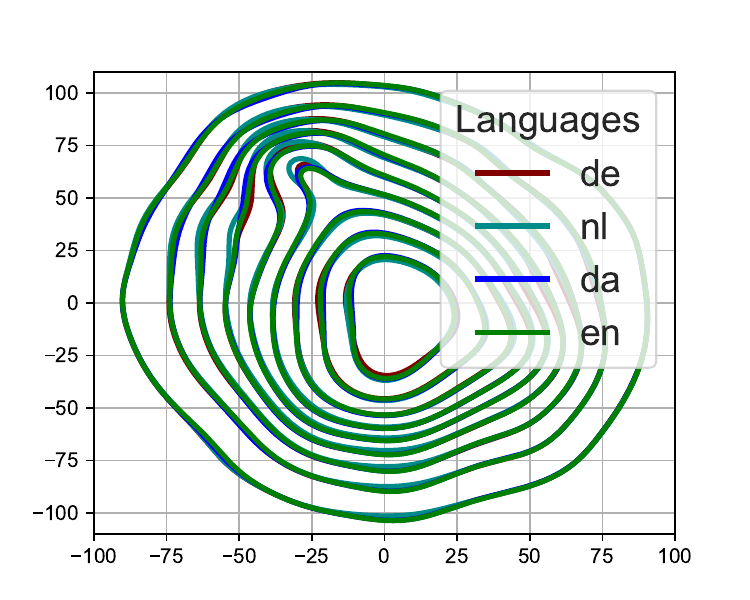}
        \caption{Germanic}
        \label{fig:append_alignment_1}
    \end{subfigure}
    \begin{subfigure}[b]{0.45\linewidth}
        \centering
        \includegraphics[width=\linewidth]{figures/alignments/representation_alignment_romance.pdf}
        \caption{Romance}
        \label{fig:append_alignment_2}
    \end{subfigure}
    \begin{subfigure}[b]{0.45\linewidth}
        \centering
        \includegraphics[width=\linewidth]{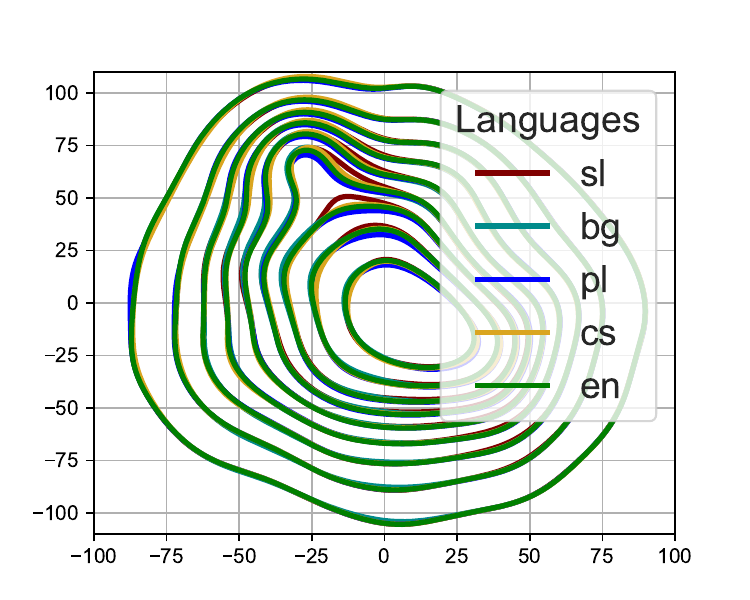}
        \caption{Slavic}
        \label{fig:append_alignment_3}
    \end{subfigure}
    \begin{subfigure}[b]{0.45\linewidth}
        \centering
        \includegraphics[width=\linewidth]{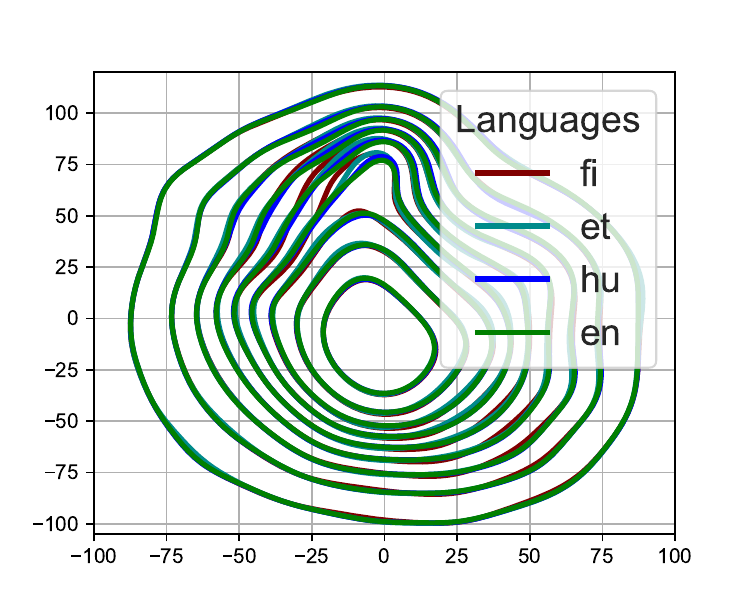}
        \caption{Uralic}
        \label{fig:append_alignment_4}
    \end{subfigure}
    \caption{Visualizations by t-SNE and BiKDE of aligning representations between \texttt{en}$\to$\texttt{en} and \texttt{x}$\to$\texttt{en} of Europarl-15 at the output of the encoder.}
    \label{fig:append_alignment}
\end{figure}
In fact, Figure \ref{fig:cluster} corresponds to the last subfigure of Figure \ref{fig:append_cluster} to show the linguistic affinity between translations from English to other languages, denoted by \texttt{en}$\to$\texttt{x}.
Specifically, Figure \ref{fig:append_cluster} shows the layer-wise states of the encoder, and Figure \ref{fig:cluster} (Figure \ref{fig:append_cluster_6}) demonstrates the state at the output of the encoder.
We employ \textit{sklearn.manifold.SpectralEmbedding}, referring to \url{https://scikit-learn.org}, to visualize the similarities computed by SVCCA (Appendix \ref{appendix:svcca}) for every layer in the encoder.
Then, we can find that representations at all encoder layers have certain clusters influenced by the families of the target languages, and the clusters become more distinct as the depth of the encoder layers increases.
This suggests that the transfer of representations to the target language begins as early as the first layer of the encoder, with gradual strengthening through further layers.
Meanwhile, this finding, i.e., even the initial encoder layers capture target language features, complements prior works \cite{investigateMNMT, lsls-2023}.

On the other hand, we follow \citet{Constras-2021} and \citet{Regular-2023} to measure the alignment of encoder representations between the identity of \texttt{en} and source languages from different families to English using t-distributed stochastic neighbor embedding (t-SNE) \cite{tsne} and bivariate kernel density estimation (BiKDE) \cite{biKDE}.
As shown in Figure \ref{fig:append_alignment}, representations from the four language families are all highly aligned with the identity pair of \texttt{en}$\to$\texttt{en}, where the common feature of those translations is the parallel semantics.
Thus, this proves that the encoder semantically aligns different translations.
However, the deep discussion should be referred to Section \ref{section:transfer}.

\section{Selecting Hyper-Parameters}\label{appendix:ablation}
\begin{figure}[t]
    \centering
        \begin{subfigure}[b]{0.44\linewidth}
            \centering
            \includegraphics[width=\linewidth]{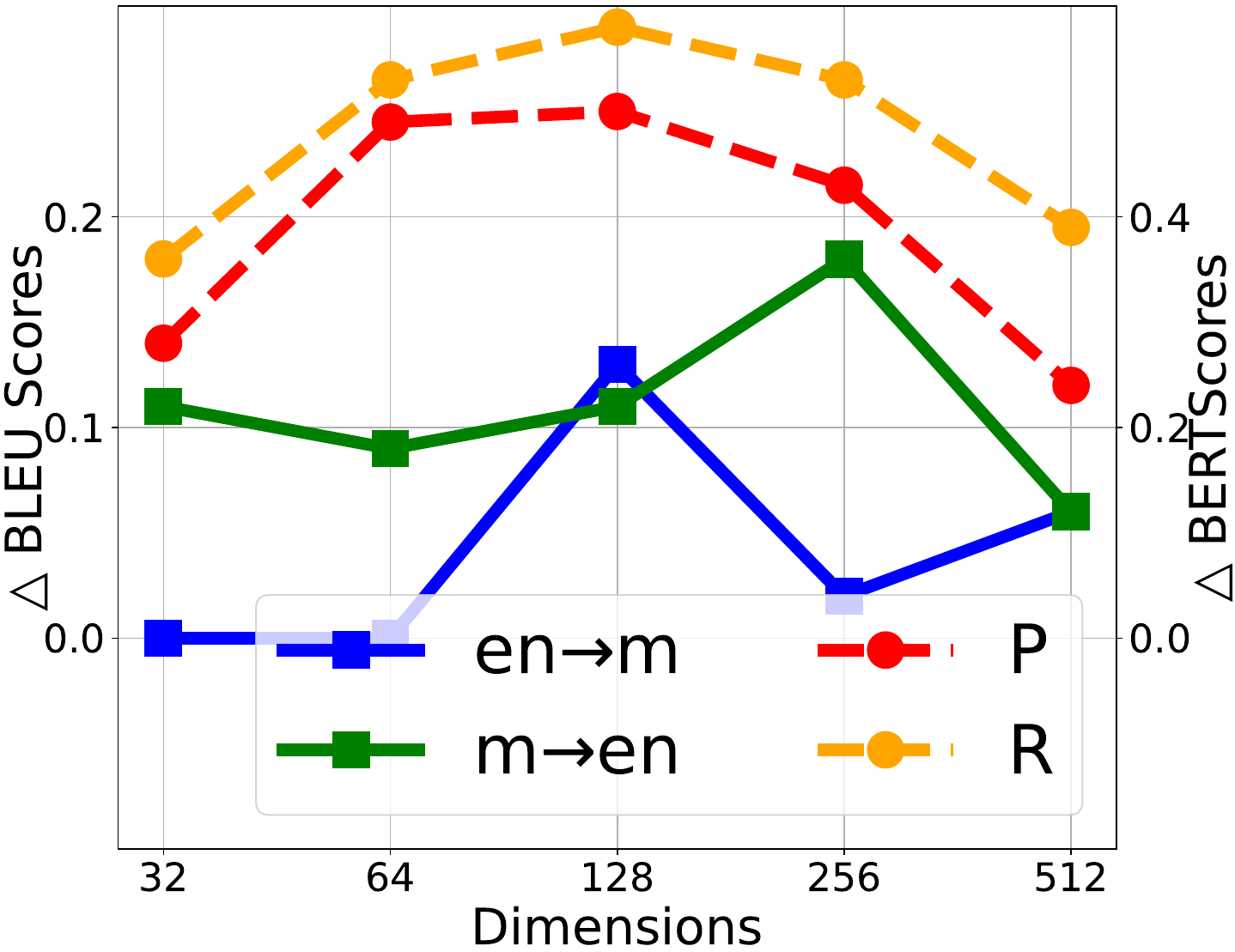}
            \caption{Dims of LoLE}
            \label{fig:ablation_1}
        \end{subfigure}
        \begin{subfigure}[b]{0.45\linewidth}
            \centering
            \includegraphics[width=\linewidth]{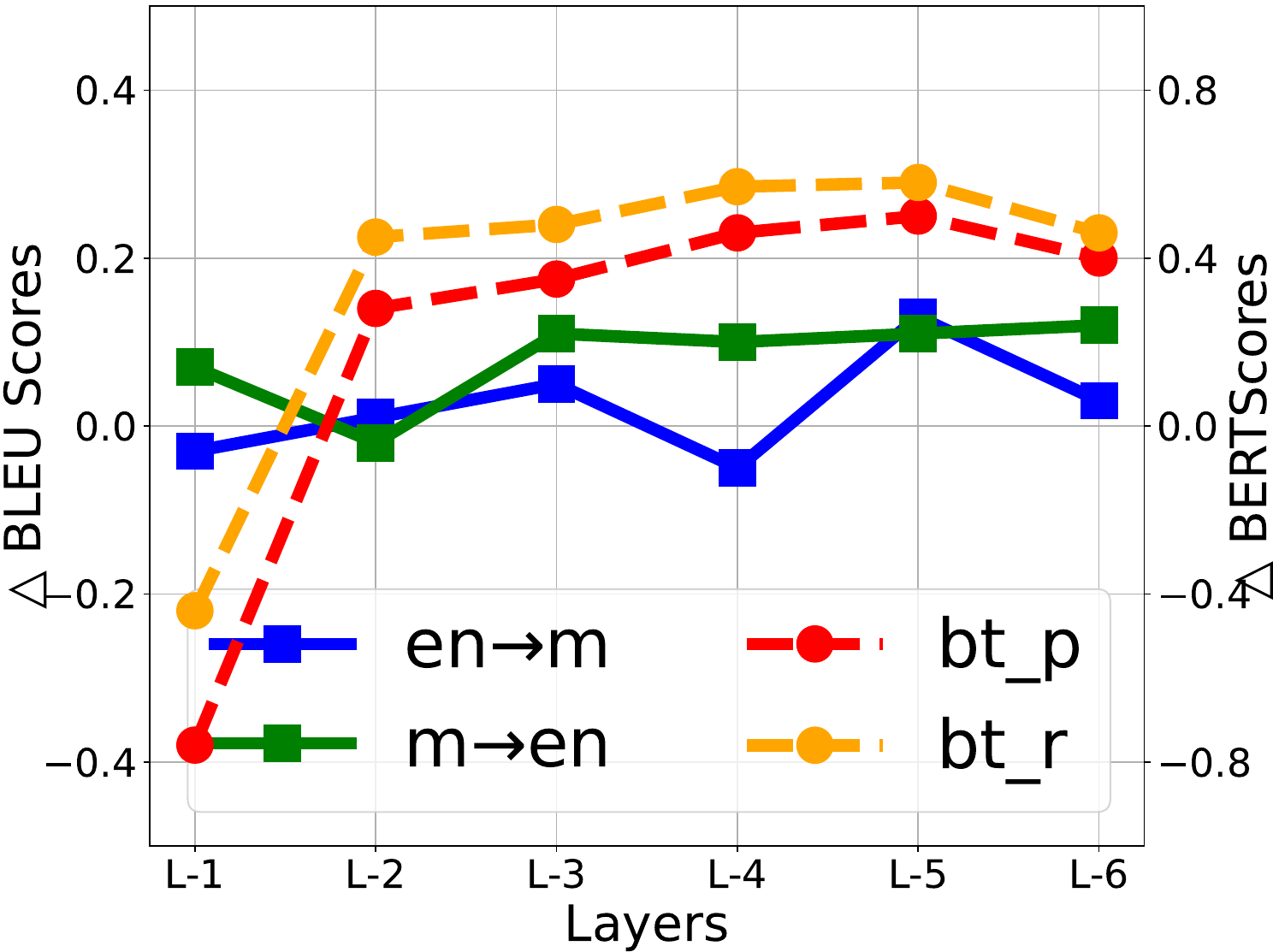}
            \caption{Layers of LoLE}
            \label{fig:ablation_2}
        \end{subfigure}
        \begin{subfigure}[b]{0.45\linewidth}
        \centering
        \includegraphics[width=\linewidth]{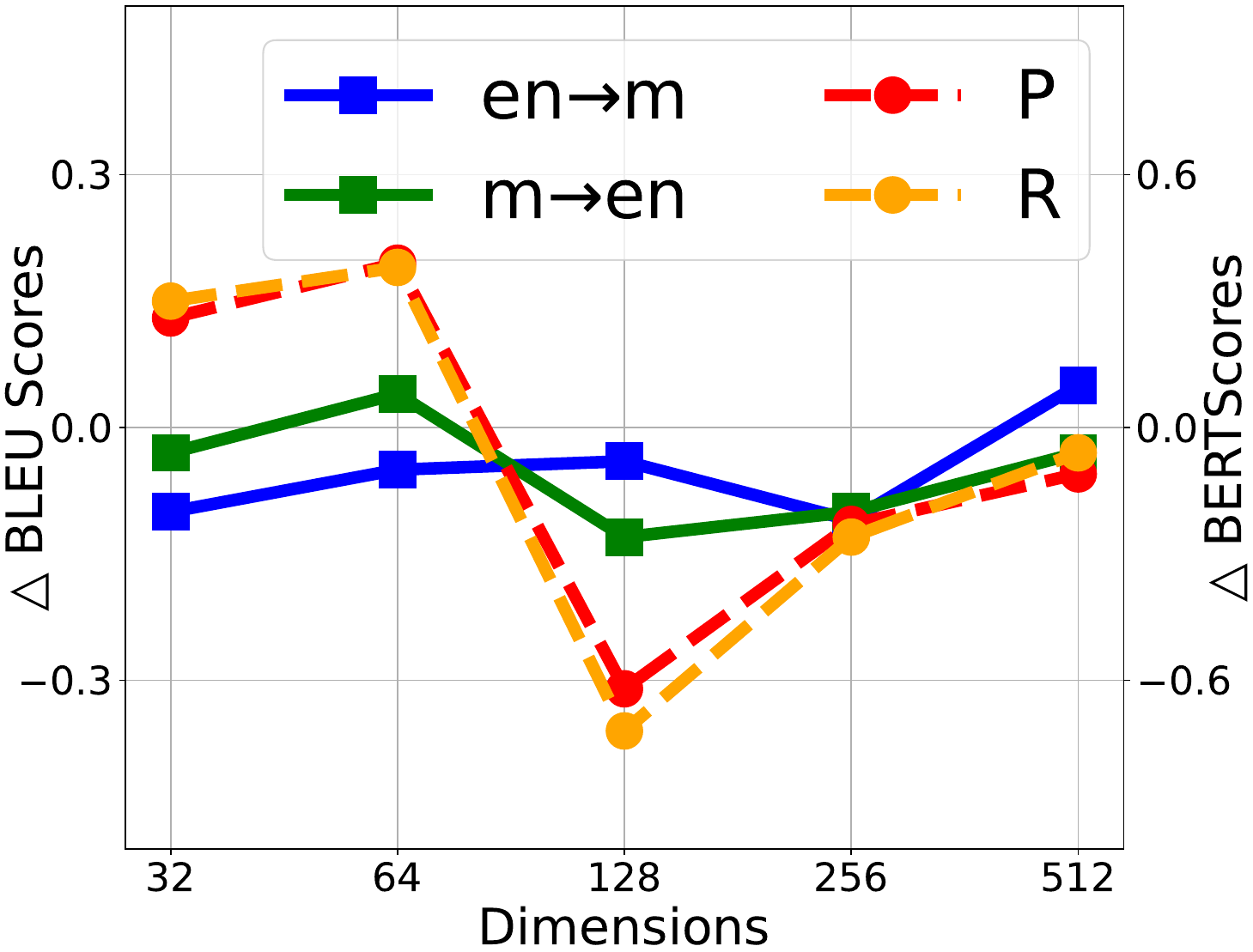}
        \caption{Dims of \textsc{LCLR} }
        \label{fig:ablation_3}
    \end{subfigure}
    \begin{subfigure}[b]{0.45\linewidth}
        \centering
        \includegraphics[width=\linewidth]{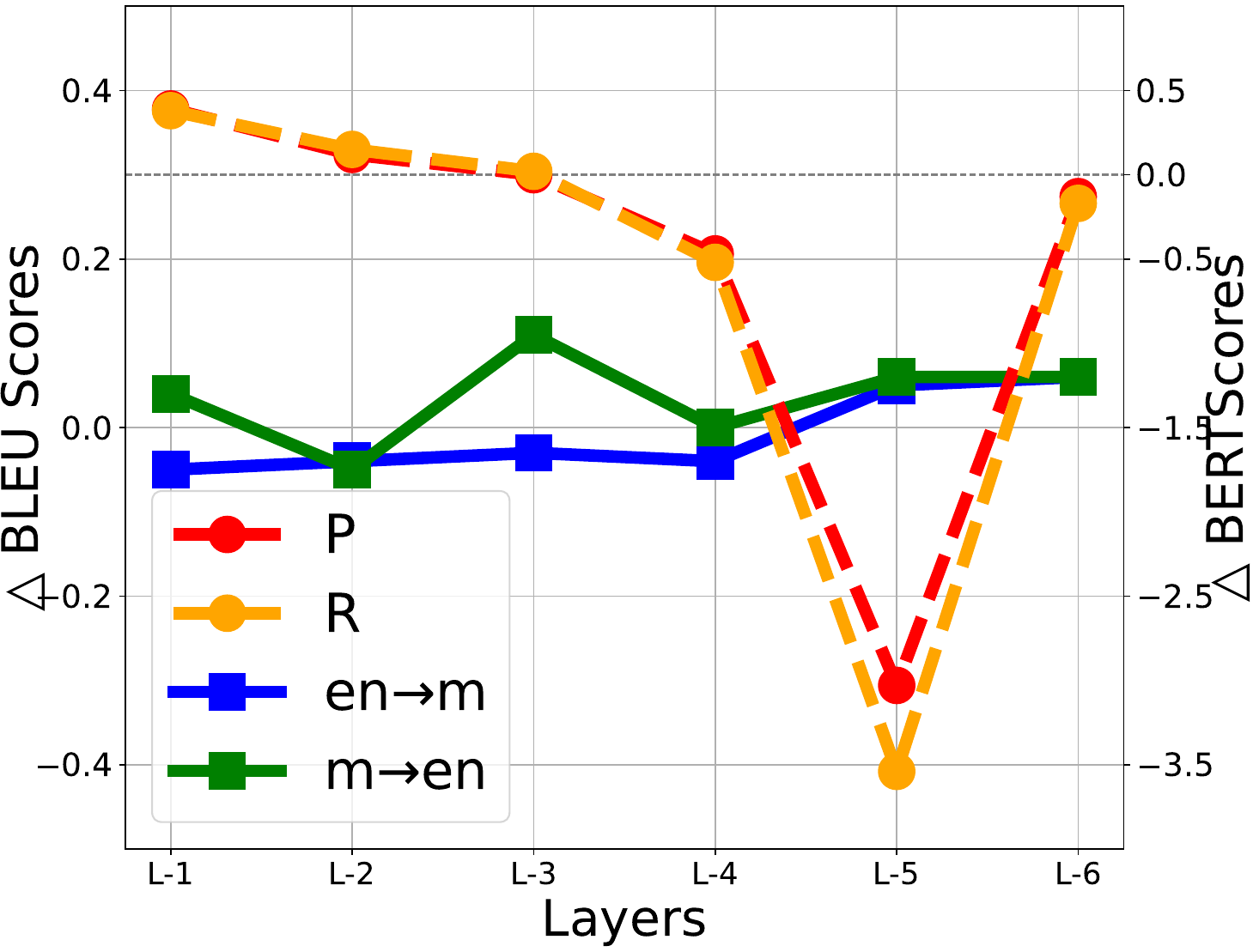}
        \caption{Layers of \textsc{LCLR} }
        \label{fig:ablation_4}
    \end{subfigure}
    \begin{subfigure}[b]{0.45\linewidth}
        \centering
        \includegraphics[width=\linewidth]{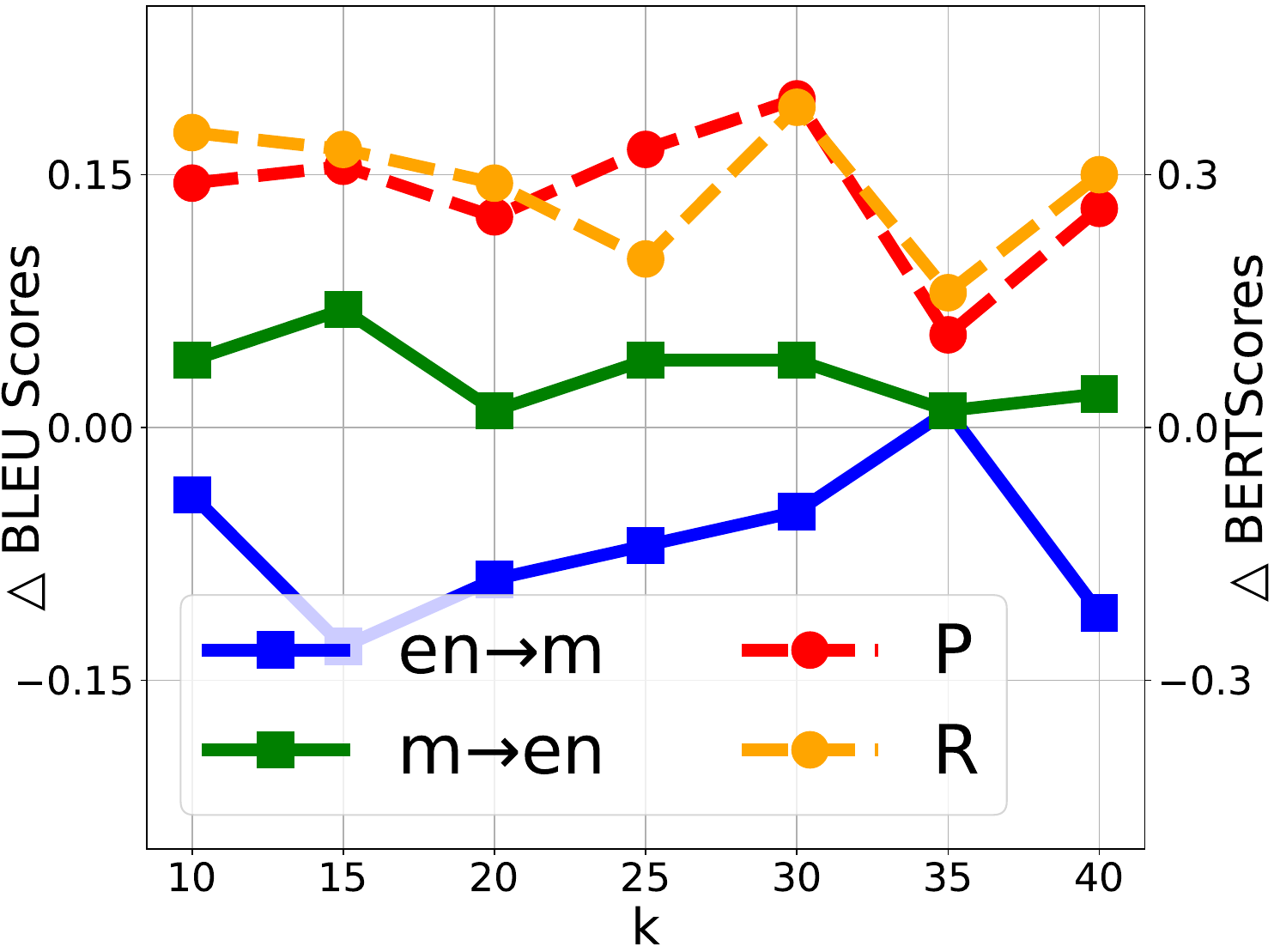}
        \caption{$k$ of \textsc{LCLR} }
        \label{fig:ablation_5}
    \end{subfigure}
    \caption{Illustrations for the ablation study. $\Delta$ means the difference between the scores of our methods and the scores of \textsc{Vanilla}. \ref{fig:ablation_1} and \ref{fig:ablation_2} present variations of \textsc{LoLE} in dimensions and layers, respectively; and \ref{fig:ablation_3}, \ref{fig:ablation_4}, and \ref{fig:ablation_5} present variations of \textsc{LCLR} in dimensions, layers, and $k$, respectively.}
    \label{fig:ablation}
\end{figure}

We conduct ablation studies on the validation set of Europarl-15 to select hyper-parameters for \textsc{LoLE} and \textsc{LCLR}, which are used in Section \ref{section:setup}.
Figure \ref{fig:ablation_1} shows that \textsc{LoLE} performs optimally with the dimension of 128, which corroborates our hypothesis in Section \ref{section:4.1}.
Figure \ref{fig:ablation_2} indicates that \textsc{LoLE} performs the best at the fifth layer and degrades significantly at lower layers, which aligns with our assertion in Section \ref{section:transfer} that lower layers of the encoder are more correlated with the source language, and enhances the language transfer benefits of transferability (Section \ref{section:4.1}).
Figure \ref{fig:ablation_3} is consistent with the theory of contrastive learning in which full dimensions lead to collapse \cite{collapse}.
Figure \ref{fig:ablation_4} demonstrates that, as the position constructed by \textsc{LCLR} increases, the scores decrease, which lends support to our analysis in Sections \ref{section:decoder} that the instability of decoder representations primarily manifests at lower layers, which also explains the weakness of \textsc{TLP} because improving the capacity of distinguishing languages is redundant for the decoder's top layer.
We also conduct an ablation study for hyperparameter $k$ for \textsc{LCLR} with a dimension of 64 at the bottom decoder layer. The results are shown in Figure \ref{fig:ablation_5}, with an empirically optimal $k=30$.

\section{Analysis of Improved Representation for Fine-tuning Pre-trained Models}\label{appendix:improve}
\begin{table}[t]
  \centering
  \resizebox{0.485\textwidth}{!}{
    \begin{tabular}{ccccccc}
    \toprule
     & Pairs & Model & Method & (\romannumeral1) & (\romannumeral2) & (\romannumeral3) \\
    \midrule
    \multirow{8}{*}{\begin{minipage}{1.2cm}\centering Encoder Side \end{minipage}} & \multirow{4}{*}{\begin{minipage}{1.3cm}\ding{172} of \texttt{zh} \\ \ding{173} of \texttt{ar}\end{minipage}} & \multirow{2}{*}{M2M} & \textsc{F.T.} & 79.66 & 100.0 & 79.66 \\
    & & & \textsc{LoLE} & 79.52 & 98.75 & 77.76 \\
    \cdashline{3-7}
    & & \multirow{2}{*}{mBART} & \textsc{F.T.} & 63.97 & 100.0 & 63.97 \\
    & & & \textsc{LoLE} & 63.52 & 97.90 & 61.66 \\
    \cdashline{2-7}
    & \multirow{4}{*}{\begin{minipage}{1.3cm}\ding{172} of \texttt{he} \\ \ding{173} of \texttt{vi}\end{minipage}} & \multirow{2}{*}{M2M} & \textsc{F.T.} & 81.50 & 100.0 & 81.50 \\
    & & & \textsc{LoLE} & 80.54 & 98.27 & 79.88 \\
    \cdashline{3-7}
    & & \multirow{2}{*}{mBART} & \textsc{F.T.} & 69.17 & 100.0 & 69.17 \\
    & & & \textsc{LoLE} & 70.46 & 97.61 & 67.04 \\
    \midrule
     \multirow{4}{*}{\begin{minipage}{1.2cm}\centering Decoder Side \end{minipage}} & \multirow{4}{*}{\begin{minipage}{1.3cm}\ding{172} of \texttt{ja} \\ \ding{173} of \texttt{he}\end{minipage}} & \multirow{2}{*}{M2M} & \textsc{F.T.} & 99.80 & 92.01 & 92.65 \\
    & & & \textsc{LoLE} & 99.73 & 89.81 & 90.66 \\
    \cdashline{3-7}
    & & \multirow{2}{*}{mBART} & \textsc{F.T.} & 98.53 & 90.76 & 89.62 \\
    & & & \textsc{LoLE} & 98.64 & 90.07 & 88.49 \\
    \bottomrule
    \end{tabular}
  }
  \caption{SVCCA scores. Each score times 100 for a clear illustration.
 (\romannumeral1) compares the identity of \ding{172} and \ding{172}$\to$\ding{173}, 
 (\romannumeral2) compares the identity of \ding{172} and \ding{173}$\to$\ding{172}, and (\romannumeral3) compares identities of \ding{172} and \ding{173}.
 Encoder Side means computing the output of the encoder, and Decoder Side means computing the output of the 1st layer of the decoder.}
  \label{tab:3}
\end{table}
Section \ref{section:improve} is the representational analysis for models, which are trained from scratch with proposed \textsc{LOLE} and \textsc{LCLR}.
We also show the representational analysis for fine-tuning pre-trained models.

Given the positive correlation shown in Section \ref{section:correlation}, we compute SVCCA scores in the same way as done in Section \ref{section:transfer} and show the results in Table \ref{tab:3}.
Unlike Section \ref{section:transfer}, we equally consider the encoder and decoder because the encoder is only related to the source language and does not transfer representations to the target language in the training strategy of M2M \cite{m2m} and mBART50 \cite{mbart50}.
Additionally, the different training strategy is the primary reason that \textsc{F.T.} shows the same scores in (\romannumeral1) and (\romannumeral3) and keeps 100.0 in (\romannumeral2). 
Alternatively, although the scores of (\romannumeral1), which reflect target language features, decrease in our methods, the scores of (\romannumeral2) and (\romannumeral3) also decrease.
As a result, the differences between the scores of (\romannumeral1), (\romannumeral2), and (\romannumeral3) increase, that is, the relative importance of target language features increases. 
This result proves our statements in Sections \ref{section:entangle} and \ref{section:correlation} again that target language features are consistently beneficial in the encoder.
On the other hand, the decoder side shows the same tendency as the encoder side.
This fits our motivation in Section \ref{section:4.2} to further improve the discriminating ability of lower layers of the decoder, although the training strategy of M2M and mBART50 has already provided a high capacity in discrimination for the decoder.

\section{Token-level Alignments in Other Cases}\label{appendix:alignment}
\begin{figure}[t]
    \centering
      \begin{subfigure}[b]{0.9\linewidth}
        \centering
        \includegraphics[width=\linewidth]{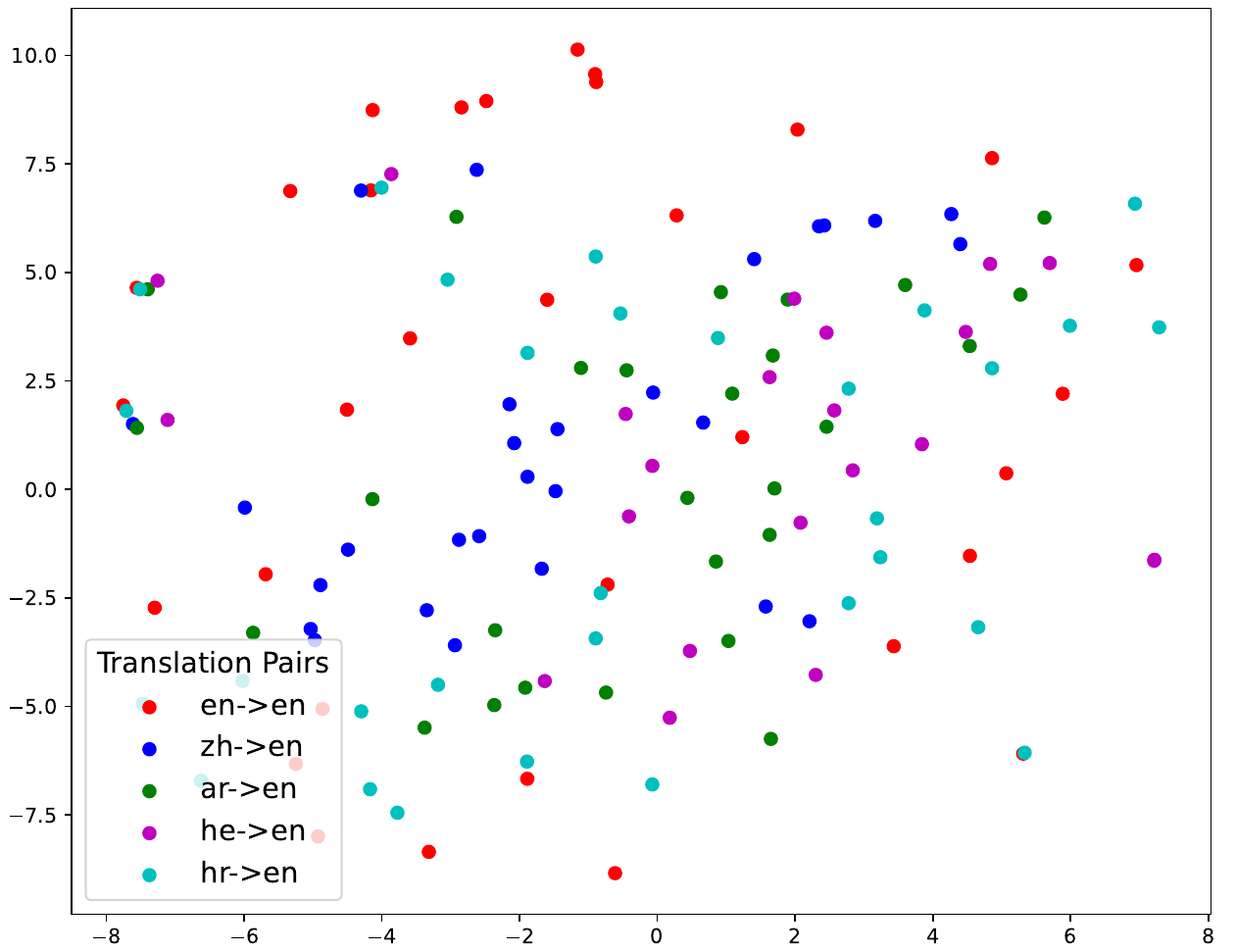}
        \caption{Semantic alignments on embeddings}
        \label{fig:append_align_1}
      \end{subfigure}
      \begin{subfigure}[b]{0.9\linewidth}
        \centering
        \includegraphics[width=\linewidth]{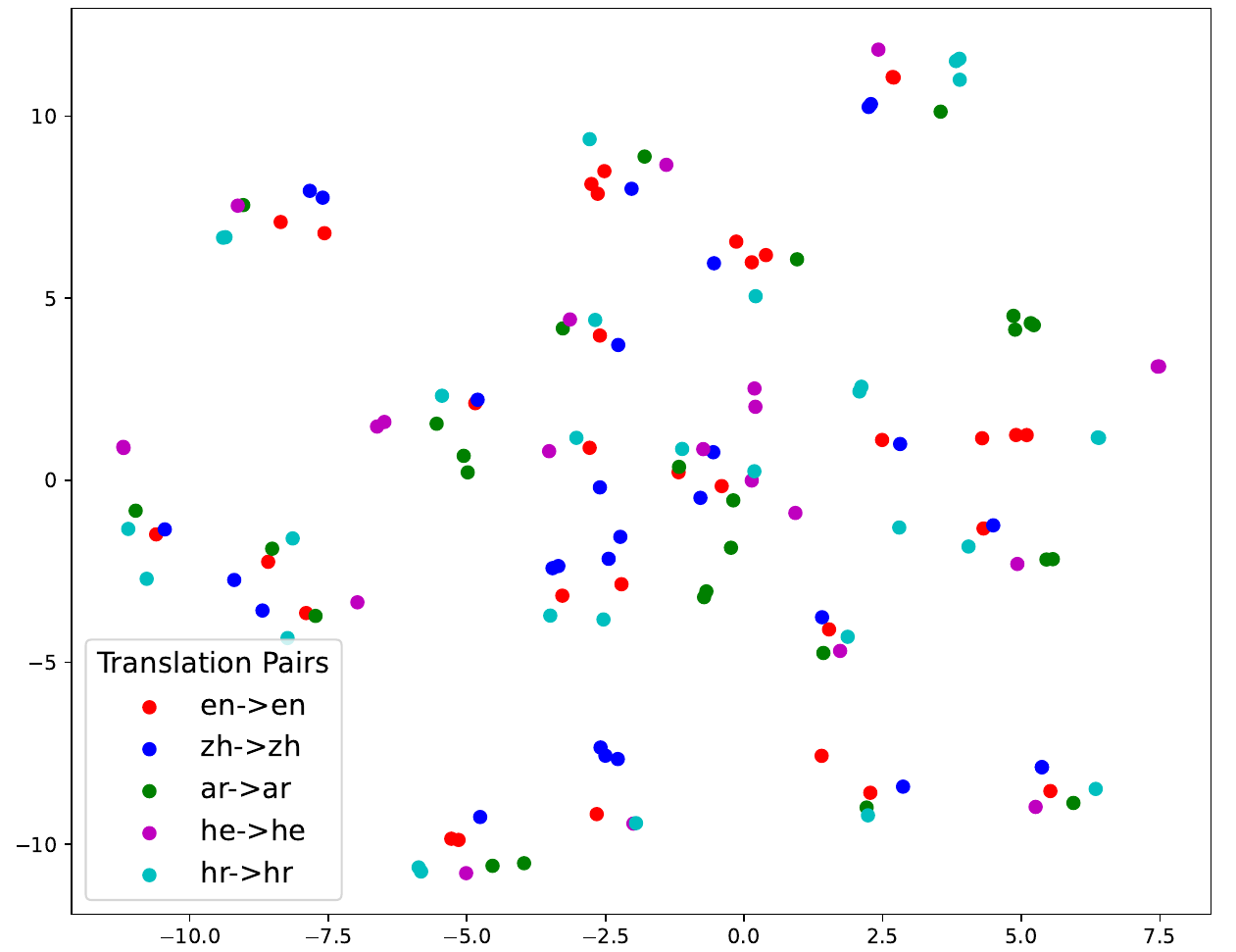}
        \caption{Semantic alignments on identities}
        \label{fig:append_align_2}
      \end{subfigure}
    \caption{Illustration of the token-level alignment corresponding to Figure \ref{fig:align}.
    Representations shown in \ref{fig:append_align_1} are collected at the embedding layer, whose overall variance is 1.45.
    Representations shown in \ref{fig:append_align_2} are collected from identities, whose overall variance is 0.13.
    }
    \label{fig:append_align}
\end{figure}
First of all, the English sentence for semantic analysis in Figures \ref{fig:word_aligned} and \ref{fig:append_align} is: By the end of this year, there will be nearly a billion people on this planet that actively use social networking sites.
Compared with the discussion in Section \ref{section:transfer}, token-level representations are not aligned at the embedding layer, and are relatively aligned in the case of using the identity pairs, where the degree of divergence is substantially higher than the case of Figure \ref{fig:word_aligned}.

\section{Evaluation Metrics Selection}\label{appendix:metric}
In this work, we select two main automatic evaluation metrics and a secondary statistic measurement.
The first one is SacreBLEU \cite{sacrebleu} which is an implementation of BLEU \cite{bleu}.
This is the most popular and common metric used in evaluating the alignment between inferences and references at the word level.
In order to counter the insufficiency of SacreBLEU, we also select BERTScore \cite{bertscore}, which is a representational metric to evaluate the semantic similarity between inferences and references.
Furthermore, to show whether the improvements brought by proposed methods are significant, we also conduct the statistical significance testing \cite{significance} using paired bootstrap resampling with 1,000 iterations and 0.5 resampling ratios, consequently, the case of $p<$ 0.05 means that the difference is significant.

Additionally, we follow prior works \cite{TLP-2021, target-off} to report the off-target ratio, which is measured by \textit{fasttext-langdetect}\footnote{\url{https://pypi.org/project/fasttext-langdetect}}.
The off-target translation refers to a sentence translated to an incorrect target language rather than the target language we expected.
However, the off-target ratio is not reliable, because the popular tools used in measuring off-target ratios are based on word level and lack support in low-resource languages.
Furthermore, the score of SacreBLEU can directly show the problem of off-target, because the evaluation process of SacreBLEU tends to give a great penalty on an inference, which has a different writing script from the expected target language.
Therefore, we only report it as a secondary metric in Appendix \ref{appendix:offratio}.

\section{Off-Target Ratio of Results}\label{appendix:offratio}
\begin{table}[t]
  \centering
  \resizebox{0.485\textwidth}{!}{
    \begin{tabular}{lcccccc}
    \toprule
    & \multicolumn{2}{c}{Europarl-15} & \multicolumn{2}{c}{TED-19} & \multicolumn{2}{c}{OPUS-100} \\
    \midrule
    Method & zero.($\uparrow$) & off.($\downarrow$) & zero.($\uparrow$) & off.($\downarrow$) & zero.($\uparrow$) & off.($\downarrow$) \\
    \midrule
    \textsc{Vanilla} & 24.65 & 1.34 & 11.98 & 4.08 & 5.04 & 70.41 \\ 
    \textsc{DisPos}  & 25.89 & 0.84 & 12.80 & 3.82 & 5.58 & 61.65 \\
    \textsc{TLP} & 24.96 & 1.22 & 12.74 & 3.71 & 4.60 & 83.29 \\
    \textsc{SemAli} & 25.25 & 0.99  & 13.45 & 3.62 & 6.42 & 58.25 \\
    \midrule
    \textsc{LoLE} & 26.09 & 0.71 & 13.20 & 3.69 & 7.92 & 50.05 \\
    \textsc{LCLR} & 25.71 & 0.79 & 12.12 & 3.86 & 5.11 & 68.53 \\
    \textsc{BOTH} & 26.20 & 0.74 & 13.31 & 3.69 & 7.97 & 55.06 \\
    \bottomrule
    \end{tabular}
  }
  \caption{
  Off-target ratio corresponding to experimental results in Table \ref{tab:1}.
  zero. indicates the BLEU scores of zero-shot translations.
  off. indicates the off-target ratio counted by all zero-shot translation pairs.
  }
  \label{tab:4}
\end{table}

\begin{table}[t]
  \centering
  \resizebox{0.485\textwidth}{!}{
    \begin{tabular}{ccccccc}
    \toprule
    Model& Metric & \textsc{Pre.} & \textsc{F.T.} & \textsc{LOLE} & \textsc{LCLR} & \textsc{BOTH} \\
    \midrule
    \multirow{2}{*}{M2M-418M}& zero.($\uparrow$) & 14.51 & 17.46 & 17.52 & 17.65 & 17.68  \\ 
    & off.($\downarrow$)  & 3.66 & 3.34 & 3.32 & 3.24 & 3.33 \\
    \midrule
    \multirow{2}{*}{M2M-418M} & zero.($\uparrow$) & 15.95 & 18.48 & 18.67 & 18.64 & 18.69  \\ 
    &off.($\downarrow$)  & 3.50 & 3.15 & 3.16 & 3.16 & 3.14 \\
    \midrule
    \multirow{2}{*}{mBART50}&zero.($\uparrow$) & 6.92 & 5.58 & 7.28 & 9.69 & 9.55  \\ 
    &off.($\downarrow$)  & 43.56 & 65.26 & 40.76 & 38.24 & 35.28 \\
    
\bottomrule
\end{tabular}
  }
  \caption{
  Off-target ratio corresponding to experimental results in Table \ref{tab:2}.
  Abbreviations follow Table \ref{tab:2} and \textsc{Pre.} refers to the model without any fine-tuning.
  In addition, compared to Table \ref{tab:4}, we switched the horizontal and vertical axes, because there is only one dataset, TED-19, used in fine-tuning experiments.
  }
  \label{tab:5}
\end{table}

Tables \ref{tab:4} and \ref{tab:5} show the measurement of the off-target ratio, which are the supplement of Tables \ref{tab:1} and \ref{tab:2}.
We can observe that the off-target ratio is always inversely proportional to BLEU scores, aligning with our discussion in Appendix \ref{appendix:metric}.
Additionally, there are two points worth noting:
(1) In Table \ref{tab:4}, the off-target ratio in OPUS-100 is generally higher.
This is not an outlier because resulting in a strong zero-shot translation capability in OPUS-100 is particularly challenging due to the large number of languages involved and the limited corpus for individual languages \cite{massive-2020, TLP-2021}.
(2) In Table \ref{tab:5}, the off-target ratio counted from mBART50 is higher than other cases.
This abnormal value has been discussed in Section \ref{section:results}, that is, the zero-shot ability of mBART50 is weaker than M2M models, and then, the fine-tuning dramatically changes the behavior of the model.

\end{document}